\documentclass[a4paper,onesided,12pt]{report}

\usepackage{fbe_tez}
\usepackage[utf8]{inputenc} % To use Unicode (e.g. Turkish) characters
\usepackage{enumitem}
\usepackage{amsmath, amsthm, amssymb}
\usepackage[bottom]{footmisc}
\usepackage{cite}
\usepackage{graphicx}
\usepackage{subfig}
\usepackage{adjustbox}
\usepackage{longtable}
\usepackage{verbatim}
\usepackage{tcolorbox}
\usepackage{booktabs}
\usepackage{lscape}

\graphicspath{{figures/}} % Graphics will be here

\usepackage{multirow}
\usepackage{algorithm}
\usepackage{algorithmic}
\usepackage{lipsum}

\title{A MULTIMODAL APPROACH FOR AUTOMATIC MANIA ASSESSMENT IN BIPOLAR DISORDER}
\turkcebaslik{Bipolar Bozuklukta Otomatik Mani Değerlendirmesi İçin Çokkipli Bir Yaklaşım}
\degree{B.S., Computer Engineering, Boğaziçi University, 2018}
\author{Pınar Baki}
\program{Computer Engineering}
\subyear{2020}

\supervisor{Prof. Lale Akarun}
\cosuperi{Prof. Albert Ali Salah}
\examineri{Assoc. Prof. Arzucan Özgür}
\examinerii{Assist. Prof. İnci Ayhan}
\examineriii{Assist. Prof. Heysem Kaya}
\dateofapproval{09.12.2020}

\begin{document}

\pagenumbering{roman}
\makemstitle % M.S. thesis
%\makeapprovalpage
\begin{acknowledgements}
First of all, I would like to thank Albert Ali Salah for always inspiring me, and encouraging me to push my boundaries. Thank you for making me feel as a part of the team. I would also like to thank Heysem Kaya for helping me in every stage of this thesis. I couldn't finish this thesis without your support and your ideas.

Furthermore, I would like to thank Elvan Çiftçi, and Hüseyin Güleç for providing me valuable insights, and I would like to thank Lale Akarun, Arzucan Özgür and İnci Ayhan for being a part of my thesis process.

Last but not least, I would like to thank my mom, Fatma Baki for always keeping my excitement levels high, my father, Zafer Baki for his endless support throughout my studies, and my sister, Yeşim Baki for being my quarantine buddy and preparing me the best snacks while I was writing this thesis.

\end{acknowledgements}
\begin{abstract}

Bipolar disorder is a mental health disorder that causes mood swings that range from depression to mania. Diagnosis of bipolar disorder is usually done based on patient interviews, and reports obtained from the caregivers of the patients. Subsequently, the diagnosis depends on the experience of the expert, and it is possible to have confusions of the disorder with other mental disorders. Automated processes in the diagnosis of bipolar disorder can help providing quantitative indicators, and allow easier observations of the patients for longer periods. Furthermore, the need for remote treatment and diagnosis became especially important during the COVID-19 pandemic. In this thesis, we create a multimodal decision system based on recordings of the patient in acoustic, linguistic, and visual modalities. The system is trained on the Bipolar Disorder corpus. Comprehensive analysis of unimodal and multimodal systems, as well as various fusion techniques are performed. Besides processing entire patient sessions using unimodal features, a task-level investigation of the clips is studied. Using acoustic, linguistic, and visual features in a multimodal fusion system, we achieved a 64.8\% unweighted average recall score, which improves the state-of-the-art performance achieved on this dataset.

\end{abstract}
\begin{ozet}
Bipolar bozukluk, depresiften manik hale varan bir erimde değişimlere neden olan bir akıl sağlığı bozukluğudur. Bipolar bozukluğun teşhisi genellikle hasta görüşmeleri ve hastaların bakıcılarından alınan raporlara göre yapılır. Hastalığın tanısı, uzmanların deneyimine bağlıdır ve hastalığın diğer ruhsal bozukluklarla karıştırılması mümkündür. Bipolar bozukluğun teşhisinde otomatik süreçler kullanılması, sayısal göstergeler sağla-
maya yardımcı olabilir ve hastaların daha uzun süreler için daha kolay gözlemlenmesini sağlar. Öte yandan, uzaktan tedavi ve teşhis ihtiyacı COVID-19 salgını sırasında özellikle önemli hale gelmiştir. Bu tezde, hastanın akustik, dilbilimsel ve görsel modalitelerde kayıtlarına dayanan çokkipli bir karar sistemi oluşturduk. Sistem, Bipolar Disorder veri seti üzerinde eğitilmiştir. Tekkipli ve çokkipli sistemlerin kapsamlı analizinin yanı sıra çeşitli füzyon teknikleri de incelenmiştir. Tüm hasta seanslarını tekkipli özellikleri kullanarak işlemenin yanı sıra, kliplerin görev düzeyindeki performansları da incelenmiştir. Çokkipli bir füzyon sisteminde akustik, dilbilimsel ve görsel özellikleri kullanarak, \%64.8 ağırlıksız ortalama geri çağırma puanı elde ettik, ve bu sonuç, şimdiye kadar Bipolar Disorder veri setinin test kümesinde elde edilen en yüksek skordur.

\end{ozet}
\tableofcontents
{%
\let\oldnumberline\numberline%
\renewcommand{\numberline}{\figurename~\oldnumberline}%
\listoffigures%
}
{%
\let\oldnumberline\numberline%
\renewcommand{\numberline}{\tablename~\oldnumberline}%
\listoftables%
}
\begin{symbols}
% The title will be typeset as "LIST OF SYMBOLS".
%
% Use a separate \sym command for each symbols definition.
% First Latin symbols in alphabetical order

%\sym{$a_{ij}$}{Description of $a_{ij}$}
%\sym{$\mathbf{A}$}{State transition matrix of a hidden Markov model}
\sym{$B(\alpha)$}{Normalizing factor in Dirichlet distribution}
\sym{$C$}{Regularization coefficient}
\sym{$\mathbf{H}$}{Hidden layer matrix}
\sym{$\mathbf{I}$}{Identity matrix}
\sym{$\mathbf{K}$}{Kernel}
\sym{$\mathbf{L}$}{Vector that contains labels}
\sym{$N$}{Order of vectors in Dirichlet distribution}
\sym{$\mathbf{P}$}{Matrix that contains class probabilities}
\sym{$\mathbf{T}$}{Output layer matrix}
\sym{$Y_t$}{YMRS score}
% Then Greek symbols in alphabetical order
\sym{$\alpha$}{Elements of the vector drawn from the Dirichlet distribution}
\sym{$\beta$}{Weight matrix between the hidden layer and the output layer}
%\sym{$\beta_t(i)$}{Backward variable}
%\sym{$\Theta$}{Parameter set}

\end{symbols}

\begin{abbreviations}
 % Abbreviations in alphabetical order
 \sym{ACII}{Asian Conference on Affective Computing and Intelligent Interaction}
\sym{ASR}{Automatic Speech Recognition}
\sym{AUC}{Area Under Curve}
\sym{AVEC}{Audio/Visual Emotion Challenge}
\sym{BD}{Bipolar Disorder}
\sym{Bi-LSTM}{Bidirectional Long Short-Term Memory}
\sym{BoAW}{Bag-of-Acoustic-Words}
\sym{BoVW}{Bag-of-Visual-Words}
\sym{BUEMODB}{Bogazici University Emotional Database}
\sym{CAE}{Convolutional Auto-Encoder}
\sym{CapsNet}{Capsule Neural Network}
\sym{CNN}{Convolutional Neural Network}
\sym{ComPaRe}{Computational Paralinguistics Challenge}
\sym{DALY}{Disability-adjusted Life Year}
\sym{DCNN}{Deep Convolutional Neural Network}
\sym{DNN}{Deep Neural Network}
\sym{DWELM}{Deep Weighted Extreme Learning Machine}
\sym{E-DAIC}{Extended Distress Analysis Interview Corpus}
\sym{eGEMAPS}{The Extended Geneva Minimalistic Acoustic Parameter Set}
\sym{ELM}{Extreme Learning Machine}
\sym{FACS}{Facial Action Coding System}
\sym{FAU}{Facial Action Unit}
\sym{FER2013}{Facial Emotion Recognition 2013} 
\sym{GEO}{Geometric Features}
\sym{HOG-LBP}{Histograms of Oriented Gradients-Local Binary Pattern}
\sym{IS10}{INTERSPEECH 2010}
\sym{KELM}{Kernel Extreme Learning Machine}
\sym{LIWC}{Linguistic Inquiry and Word Count}
\sym{LLD}{Low Level Descriptors}
\sym{LSTM}{Long Short-Term Memory}
\sym{MADRS}{Montgomery-Asberg Depression Rating Scale}
\sym{MFCC}{Mel Frequency Cepstral Coefficient}
\sym{ML}{Machine Learning}
\sym{MLP}{Multi Layer Perceptron}
\sym{MM1}{Multimodal 1}
\sym{multi-DDAE}{Multimodal Deep Denoising Autoencoder}
\sym{MUMBAI}{the Multi-Person, Multimodal Board Game Affect and Interaction Analysis Dataset} 
\sym{NLP}{Natural Language Processing}
\sym{NLTK Vader}{Natural Language Toolkit Valence Aware Dictionary for sEntiment Reasoning}
\sym{openSMILE}{Open Speech and Music Interpretation by Large Space Extraction}
\sym{PCA}{Principal Component Analysis}
\sym{PRIORI}{Predicting Individual Outcomes for Rapid Intervention}
\sym{RBF}{Radial Basis Function}
\sym{RNN}{Recurrent Neural Network}
\sym{SIFT}{Scale-invariant Feature Transform}
\sym{SVM}{Support Vector Machine}
\sym{TF-IDF}{Term Frequency-Inverse Document Frequency}
\sym{UAR}{Unweighted Average Recall}
\sym{VGG-Face}{Visual Geometry Group-Face}
\sym{WELM}{Weighted Extreme Learning Machine}
\sym{W-KELM}{Weighted Kernel Extreme Learning Machine}
\sym{YMRS}{Young Mania Rating Scale}
\end{abbreviations}

\chapter{INTRODUCTION}
\label{chapter:introduction}
\pagenumbering{arabic}

Bipolar disorder (BD) is a mental health condition that causes extreme mood swings like emotional highs (mania, hypomania), lows (depression), mixed episodes where depression and manic symptoms occur together. The diagnosis of bipolar disorder requires lengthy observations on the patient. Otherwise, it can be mistaken with other mental disorders like anxiety or depression. The disease affects 2\% of the population, and sub-threshold forms (recurrent hypomania episodes without major depressive episodes) affect an additional 2\%~\cite{merikangas2007lifetime}. It is ranked as one of the top ten diseases of disability-adjusted life year (DALY) indicator among young adults, according to World Health Organization~\cite{world2008global}. It takes 10 years on average to diagnose bipolar disorder after the first symptoms~\cite{lish1994national}.

In bipolar disorder, the clinical appearance of the patients changes based on the moods they are in. The changes are seen in both their sound and visual appearance, as well as the energy level changes. In the manic episode, the speech of the patient becomes louder, rushed, or pressured. The patient can be very cheerful, furious, or overly confident. The movements of the patient become more active, exaggerated, and they tend to wear very colorful clothes. Feelings and the state of mind change quickly. Racing thoughts, reduced need for sleep, lack of attention, increase in targeted activity (work, school, personal life) are some situations patients can experience in the manic episode. These symptoms return to a normal state during the remission state~\cite{elvan2017}. 

Today, the diagnosis of mental health disorders rely on questionnaires done by psychiatrists and reports from patients and their caregivers. Psychiatrists perform some tests to collect information about the patient's cognitive, neurophysiological, and emotional situations~\cite{elvan2017}. But these reports are subjective, and there is a need for more systematic and objective diagnosis methods. Especially, with the COVID-19 pandemic, remote treatment and diagnosis gain importance, which can be achieved using automated methods.

One of the tools used to rate the severity of the manic episodes of a patient is the Young Mania Rating Scale (YMRS). During the interviews, psychiatrists observe the patient's symptoms and give ratings to them. The 11 items in YMRS assess the elevated mood, increased motor activity-energy, sexual interest, sleep, irritability, speech rate and amount, language-thought disorder, content, disruptive-aggressive behavior, appearance, and insight. Most of these can be observed from speech patterns, body or facial movements, and the content of what was spoken during the interview.

Recent advancements in technologies like social media, smartphones, wearable devices, and improvements in recording techniques like better cameras, neuroimaging techniques, microphones enable us to gather good quality data from people during their everyday lives. This creates an opportunity to create tools to monitor the symptoms of the patients in longer periods, screen patients before they see the psychiatrists, complement clinicians in the diagnosis, and capture their behaviors in situations where they cannot act or hide the symptoms.

In recent years, there are many works on diagnosing psychiatric disorders like Alzheimer's disease, anxiety, attention deficit hyperactivity disorder, autism spectrum disorder, depression, obsessive-compulsive disorder, bipolar disorder~\cite{shatte2019machine} using machine learning (ML) techniques. The datasets used for the detection of the diseases contain linguistic, auditory, and visual information. Adapted from real life, using the modalities together with fusion techniques improves the results as explained in Chapter \ref{chapter:relatedworks}.

\section{Problem Statement}
\label{section:problemstatement}

In this thesis, we focus on the question of how we can use the available modalities (acoustic, linguistic, visual) and machine learning techniques for diagnosing and classifying BD states. We work on the Bipolar Disorder Dataset collected by Çiftçi et al.~\cite{cciftcci2018turkish}, which contains patient interviews as video recordings recorded by the psychiatrists. Taking advantage of the dataset containing all acoustic, linguistic, and visual information, we investigate the results of all three modalities as unimodal systems, and multimodal systems using various fusion methods. We further investigate which kind of tasks performed by the patients (i.e. positive, neutral, or negative expected effect) are more effective for classifying BD states.

One of the main challenges was the small size of the BD dataset which contains 104, 60, and 54 samples for training, validation, and testing sets, respectively. Subsequently, it was challenging to create a model that generalizes well, while not overfitting the data. The BD dataset was collected from BD patients in a psychiatric hospital. Like other real-life datasets, samples are noisy, and there are sounds other than the patient's speeches, like door knocks, a speech of the doctor, and other sounds coming from outside of the room. Besides, some patients do not talk enough to make good generalization.

Naturalistic human behaviour datasets showing affective states are important for developing automatic analysis tools. In a related work, we have created one such resource during the eNTERFACE Summer workshop at Bilkent University, collected from board-game sessions where four-player plays are recorded with multiple cameras. The dataset is called the Multi-Person, Multimodal Board Game Affect and Interaction Analysis Dataset (MUMBAI)~\cite{schimmel2019mp,jmui}, and contains 62 game sessions, with 46 hours of visual materials in total. Similar to the BD dataset, the MUMBAI dataset contains multimodal social signals, which can be used to investigate the psychological situations of the participants. However, in the BD dataset patients are not interacting in a social group, so they can be investigated individually, while the MUMBAI dataset can be used to study multi-person interactions as well, as well as to determine individual properties, such as player experience, from social cues jointly~\cite{olalere21hbu}. 

The MUMBAI dataset is annotated manually with expressive moment labels like positive, negative, focused moments, and game-related emotion labels which are anxious, bored, confused, and delighted. Furthermore, self-reported personality and game experience tests are collected from each participant. For extracting face and head features, the OpenFace tool is used~\cite{baltruvsaitis2016openface}, and head movement, gaze behavior, affective facial expressions, mouth movements, categorical gaze direction, and facial action units are extracted. %Taking part in the creation of the MUMBAI dataset, the labeling process, and obtaining the baseline results helped me get better knowledge while working on the BD dataset.

\section{Contributions}
Using acoustic, textual, and visual modalities in a multimodal system, we achieved a 64,8\% unweighted average recall (UAR) score on the test set of the BD dataset, which advances the state-of-the-art result achieved on this dataset. For multimodal fusion, we worked on majority voting, feature fusion, and weighted sum methods, and showed their effectiveness on the BD dataset. For the acoustic, and linguistic modalities we proposed new feature sets and compared their results. Besides the entire clip level experiments, which was done using every clip as a single dataset, we further investigated the effect of the tasks on the classification performance separately, by grouping the same emotion eliciting tasks, and all tasks together to increase the dataset size.

A paper was submitted and accepted as a poster presentation at the ``27th Signal Processing and Communications Applications Conference" discussing some of the results of the thesis~\cite{baki20}.

\section{Structure of the Thesis}

In Chapter~\ref{chapter:relatedworks} we describe the related work on the classification of mental health disorders using various data sources, and in Section~\ref{section:mental} we talk about all the works that used the Bipolar Disorder Dataset so far. In Chapter 3, we introduce Bipolar Disorder Dataset. In Chapter 4, we explain the features we used for acoustic, linguistic, and visual modalities, the preprocessing methods, the feature selection methods, classification algorithms, and the fusion methods we use to create multimodal systems. In Chapter 5, we present the results of our experiments on the unimodal, and fusion systems, and the experiments performed on the tasks obtained from the clips. In Chapter 6, we discuss our results, contributions, and compare our work with the state-of-the-art. Finally, in Chapter 7 we conclude our work and explain future studies that can be performed on this dataset.

\chapter{RELATED WORK}
\label{chapter:relatedworks}
Assessment of mental health disorders using machine learning methods has been an active research area. Many researchers are working on recognizing mental health disorders varying from depression, Alzheimer’s disease, anxiety to bipolar disorder. The interdisciplinary research between psychiatrists and computer scientists helps to create new datasets and bringing insights from the medical domain to artificial intelligence. 

\section{Prediction of mental health disorders}
\label{section:mental}
The datasets used in the prediction of mental health disorders contain various data types~\cite{shatte2019machine}. Datasets are collected by psychiatrists like electronic health records~\cite{ojeme2016selecting}, surveys~\cite{hou2016big}, interviews~\cite{cciftcci2018turkish}, clinical assessments~\cite{brasil2009towards}, brain imaging scans~\cite{labate2014eeg} or gathered from the personal information of the patients outside the clinic like social media posts
~\cite{wang2013depression}, suicide notes~\cite{pestian2008using} or wearable sensor data~\cite{zhou2015tackling}. These datasets contain visual, auditory, textual, or biological information, which allows researchers to develop algorithms using computer vision, signal processing, speech processing, or natural language processing models. Some of the datasets are suitable for using modalities together, which is similar to the human decision making process. For instance, from the patient interviews recorded with a video and audio, visual, auditory, and textual features can be extracted. State-of-the-art results are achieved by the fusion of the modalities~\cite{girard2015automated}. 

Acoustic and visual cues are used in the detection of major depressive disorder in ~\cite{meng2013depression}. They use motion history histograms to extract dynamic features from video and audio data and represent the subtle change of emotions in depression. Decision level fusion of audio and visual modalities proves the effectiveness of the proposed model.

In ~\cite{cohn2009detecting}, facial action and vocal prosody (suprasegmental) features are extracted from patient interviews conducted by a clinical interviewer. Vocal prosody features provide information about the sound in language beyond the meaning of the language, like rhythm, stress, intonation etc. Support vector machine (SVM) is used for the classification of facial action unit features and logistic regression for the classification of the acoustic prosody features. Both modalities give promising results separately. However, the fusion of the audio-visual features wasn't performed in this paper. 

Another work on the recognition of depression applies hierarchical classifier systems to vocal prosody features and local appearance descriptors extracted from the faces of the patients. Kalman filter is used for the fusion of modalities~\cite{kachele2014fusion}. It enables the system to perform better in real-time and can deal with sensor failures. Their late fusion method cannot outperform the results obtained from the auditory and visual modalities separately. They stated that the performance gap between the audio and video modalities is the reason of the performance drop in the fusion results.

Similarly, audio and visual modalities are commonly used in the detection of bipolar disorder. One of the early works on the classification of bipolar disorder~\cite{karam2014ecologically} presents The University of Michigan Prechter Acoustic Database, which contains cellular phone recordings of BD patients. An SVM classifier with linear and radial basis function (RBF) kernel is used for the classification. 23 low-level speech descriptors are extracted with the Munich open speech and music interpretation by large space extraction (openSMILE) toolkit~\cite{eyben2010opensmile} from the phone recordings of the patients. The model differentiates between hypomania vs euthymia (healthy state) and depression vs euthymia with 0.81 and 0.67 area under curve (AUC) respectively. 

Muaremi et al.~\cite{muaremi2014assessing} collected a cellular phone dataset from the 12 bipolar patients of a psychiatric hospital. Using the openSMILE toolkit, they extract root mean square, mel-frequency cepstral coefficients (MFCC), pitch, harmonics-to-noise ratio, zero-crossing-rate, and summarize these low-level descriptors (LLDs) with 12 functionals. Besides these acoustic features, they also experiment with phone call statistics like number of phone calls during the day, average duration of the phone calls etc., and social signal processing features like average speaking length, average number of speaker turns etc. Among the three feature set, acoustic features perform the best. The highest performance is achieved with the early fusion of the three modalities. Using a random forest classifier, 83\% $F_{1}$ is achieved on 2 classes (manic vs. normal or depressive vs. normal).

Besides speech cues, motor activity related information (body movement, motor response time, level of psychomotor activity) is used for BD classification in~\cite{maxhuni2016classification}. The speech data is collected from cell phones of the BD patients. During the conversations over the phone, motor activity data is collected from the accelerometer on the phone. Information related to the motor activity is also collected from the self-assessment questionnaires regarding the patient's psychological state, physical state, and activity level. Their result suggests that the fusion of accelerometer features with the speech related features gives 82\% accuracy in the classification of a manic episode. With the information from the questionnaires, the final result is improved slightly to 85\%. However, they argue that the usage of questionnaires may harm the fully autonomous nature of the system.

In ~\cite{aldeneh2019identifying}, the dialogues in the assessment phone calls between the patient and the clinician from the Predicting Individual Outcomes for Rapid Intervention (PRIORI) dataset are investigated. A set of high-level dialogue features (floor control ratio, turn hold offset, number of consecutive turns, number of turn switches per minute, turn switch offsets, turn lengths) are extracted and summarized using mean and standard deviation. They also extract rhythm features (power distribution, rate, and rhythm stability) from the audio and calculate statistics using mean, standard deviation, kurtosis, skewness, max, min and their normalized locations, linear regression slope, intercept, and error functionals. For the classification of the euthymia vs depression and euthymia vs mania, logistic regression, SVM, and deep neural network (DNN) models are used. Experiments are performed on dialogue features, rhythm features, and their early fusion. For the depression detection, the fusion of the two sets of features improved the overall result, however for the mania detection additional dialogue features do not improve the results obtain with rhythm features.

\section{Related Works on the Bipolar Disorder Dataset}
\label{section:bd}
The Audio/Visual Emotion Challenge (AVEC) was held for the eighth time in 2018. The mission of the AVEC series is to create a common benchmark and pushing the boundaries of audio-visual emotion and health recognition problems. Some of the previous challenge topics were prediction of self-reported severity of depression, detecting discrete emotion classes, prediction of continuous-valued dimensional effect, depression analysis from human-agent interactions, and emotion recognition from human behaviors captured in-the-wild.

In the 2018 AVEC Challenge, a Bipolar Disorder (BD) corpus was made available~\cite{ringeval2018avec}. Several groups have worked on this corpus within the AVEC Challenge, where the goal was to determine the state of the patient given a short video sequence containing several pre-determined tasks~\cite{yang2018bipolar, du2018bipolar, xing2018multi, syed2018automated, ebrahim2018determine, schuller2019capsule, ren2019multi}. Our research group, as the creator of the bipolar challenge, did not run in this challenge as a participant, but provided the baseline and the protocol.

As a performance metric, the unweighted average recall (UAR) score is used during the challenge. Throughout this study, we also use UAR for presenting the results to compare our findings with the previous studies. In more detail, UAR is the unweighted average of the class-specific recalls obtained from the system for each of the three classes.
\begin{equation}
    UAR = \dfrac{1}{3} ( recall(remission) + recall(hypomania) + recall(mania) )
\end{equation}

Most of the works in the challenge extract both audio and visual features, and apply either decision or feature-level fusion~\cite{yang2018bipolar, xing2018multi, syed2018automated, ebrahim2018determine}. All of them obtain their best results using a fusion of these modalities.

\begin{table}[H]
\begin{center}
\caption[Comparison of the works that use the BD dataset.]{Comparison of the works that use the BD dataset. Validation and test set results are presented as UAR scores.}
\resizebox{\textwidth}{!}{\begin{tabular}{|c|c|c|c|c|}
\hline
\textbf{Paper}                                                                           & \textbf{Features}                        & \textbf{Classifier} & \textbf{Validation} & \textbf{Test} \\ \hline
Ringeval {\em et al.}~\cite{ringeval2018avec}         & eGEMAPS+FAUs  & SVM & 0.550                & \textbf{0.574}         \\ \hline
Yang {\em et al.}~\cite{yang2018bipolar}         & Arousal and upper body posture features  & Multistream & 0.783                & 0.407         \\ \hline
Du {\em et al.}~\cite{du2018bipolar} & MFCC                                     & IncepLSTM & 0.651                & -             \\ \hline
Xing {\em et al.}~\cite{xing2018multi}    & eGEMAPS+MFCC+FAUs+eyesight features       & Hierarchical recall model & 0.867                & \textbf{0.574}         \\ \hline
Syed, Sidorov, Marshall ~\cite{syed2018automated}                   & AUs+gaze+pose                            & GEWELMs & 0.550                & 0.482         \\ \hline
Ebrahim, Al-Ayyoub, Alsmirat ~\cite{ebrahim2018determine}   & MFCC+eGEMAPS+BoAW+DeepSpectrum+FAUs+BoVW & Bi-LSTM & 0.592                & 0.444         \\ \hline
Amiriparian {\em et al.}~\cite{schuller2019capsule}                   & Mel-Spectogram                           & CapsNet & 0.462                & 0.455         \\ \hline
Ren {\em et al.}~\cite{ren2019multi} & MFCC                                     & Multi-instance learning & 0.616                & \textbf{0.574} \\ \hline
Zhang {\em et al.}~\cite{zhang2020multimodal} & MFCC+FAUs+gaze+Paragraph Vector                                     & Deep Neural Network & 0.709                & -         \\ \hline
Abaei, Al Osman ~\cite{abaei2020hybrid} &      CNN                                & LSTM & 0.606                & \textbf{0.574}         \\ \hline
Sun {\em et al.}~\cite{sunbipolar} & \shortstack{\\MFCC+eGEMAPS+BoAW+DeepSpectrum+\\Soundnet18+ComParE+FAUs+BoVW+CAE+\\CNN+MSDF+HOGLBP+Geometric+BoTW}                                     & SVM  & 0.931                & -         \\ \hline
\end{tabular}}
\end{center}
\end{table}

Approaches to different mental health assessment problems have been inspired by each other. In \cite{stasak2016investigation}, experiments showed that arousal is more effective in depression assessment than valence and dominance. By considering this result,  \cite{yang2018bipolar} uses histogram based arousal features for the classification of BD episodes. This is one of the first works that use affective dimensions for this task. The proposed approach fuses acoustic features with a set of visual features. These visual features are histograms of displacement ranges that are extracted by taking the vertical and horizontal displacement of upper body keypoints. Intuitively, these features show how much upper body movement occurs during the session, which is a visual indicator of arousal. The audio-based arousal features (also based on histograms) give good results and demonstrate that emotion information in speech is relevant to the classification of the BD moods. Besides, they claim that classification on male clips gives better results than women clips, which suggests that males reflect the moods more clearly than women. Considering the small sample sizes (34 males and 16 females) and the limited cultural variation (all subjects are Turkish), there is no strong basis for such generalizations.

\cite{xing2018multi} uses textual features in addition to visual and audio based features. After getting translated text with the Automatic Speech Recognition tool of Google Cloud Platform, they extract linguistic features (number of words, sentences, unique words etc.)  using various Natural Language Processing tools. However, the experiments done using textual features are not mentioned further in the paper. For classification, they used a multi-layer hierarchical recall model where they classify the mania levels in different layers. For instance, at the first level, they classify the mania and remission classes, if the output probability is less than a threshold, that sample is classified in the second level. Using this model with The Extended Geneva Minimalistic Acoustic Parameter Set (eGEMAPS), MFCC, facial action units and gaze features, they achieve the highest UAR on the validation set among the challenge participants. However, the 0.867 UAR score on the development set and the 0.574 UAR score on the test set shows that the proposed model learns the training data too well but can not generalize to the unseen test set data, which is called overfitting.

Fisher vector encoding is a popular aggregation method mostly used in image classification or retrieval problems~\cite{sanchez2013image}. Recently, it has also been applied to several signal processing problems and promising results were obtained~\cite{kaya2016fusing}. \cite{syed2018automated} uses this approach with the Computational Paralinguistics ChallengE (ComParE) feature set. They propose some turbulence features that represent the sudden changes in feature contours of both audio and visual modalities. The classification is done using the Greedy Ensemble of Weighted Extreme Learning Machines (ELM)~\cite{zong2013weighted} where they train many weighted ELMs, then select the ones which have a UAR score more than a fixed threshold on the validation set. Turbulence features extracted from the visual modality achieve the best test set result of the challenge.

There are a couple of papers that use deep learning methods on the BD set. There are 218 samples from 46 individuals in the BD corpus. So, deep learning based models often cause an over-fitting problem on this corpus, and lead to significant drop of performance on the test set, compared to the performance on the validation set. 

In~\cite{du2018bipolar}, this problem is handled using L\textsubscript{1} regularization while using a network consisted of an Inception module combined with an long short-term memory (LSTM) network. 16-dimensional MFCC features are extracted from the speech files. Using only audio features, 0.651 UAR is achieved on the development set. However, no score is reported for the test set. 

In~\cite{ebrahim2018determine}, LSTM and Bidirectional LSTM (Bi-LSTM) models are trained on the baseline features provided by the challenge organizers. These features contain MFCCs, eGEMAPS, Bag-of-Acoustic-Words (BoAW), DeepSpectrum extracted from audio, and Facial Action Units (FAU) and Bag-of-Visual-Words (BoVW) extracted from the video. The fusion of MFCC and BoVW features on the Bi-LSTM model achieves 74.60\% UAR on the development set. However, the test set result is reported as 33.33\%, which is the chance level UAR score for the 3 class classification problem. This result emphasizes the importance of avoiding overfitting while using deep learning models. Their best result on the test set is achieved on the Bi-LSTM network trained on the concatenation of all the features provided in the challenge. In that scenario, 59.24\% of development set UAR, and 44.44\% test set UAR is achieved. Yet, this result is lower than the baseline results on the dataset achieved using the SVM model. This shows that using more complex deep learning models on a small dataset does not necessarily improve the performance.

Using the visual modality, the baseline test set score is achieved in  ~\cite{abaei2020hybrid}. The visual features are extracted from a pre-trained Visual Geometry Group-Face (VGG-Face) model fine-tuned with the Facial Emotion Recognition 2013 (FER2013) corpus~\cite{goodfellow2013challenges}. All layers are freezed except the last pooling layer, and a final layer is defined with 512 neurons, which gives a 512-dimensional feature vector for each frame. Finally, the extracted features are fed to an LSTM network. The proposed CNN-LSTM model achieves 60,6\% and 57,4\% on development and test sets respectively, which shows that the model does not overfit the data. 

To deal with the small size of the BD corpus, Capsule Neural Network (CapsNet)~\cite{sabour2017dynamic} is used in~\cite{schuller2019capsule}. In CapsNet the pooling layer in the Convolutional Neural Network (CNN) is changed with the capsules, which are a group of neurons that allow the model to learn spatial relationships between different parts of the data (mostly image), so different transformations of the data can be recognized without reducing the performance, which makes the model more efficient when working with small datasets. Mel-frequency spectrograms are extracted from the small segments of raw audio files to train the CapsNet model. Two more audio representation learning frameworks, namely AUDEEP and DEEPSPECTRUM, are proposed to compare the results with the CapsNet model. DEEPSPECTRUM features are evaluated using a linear SVM model and AUDEEP features are used in the training of the Multi-Layer Perceptron (MLP) model. Although they get similar results in all three models, AUDEEP features give the best test result, which is 49.8\%. As stated in the paper, due to the high computational cost needed for the optimization of the CapsNet hyperparameters, authors could not optimize the parameters enough and fully evaluate the best result that can be achieved using CapsNet. 

Another technique that can be used while classifying small datasets with deep learning models is multi-instance learning. In~\cite{ren2019multi}, audio clips are segmented into chunks to increase the dataset size. However, each clip has only one label and after segmenting the clip, each chunk becomes weakly labeled. For example, a clip may be labeled as 'mania', but a small chunk from that clip may not represent any 'mania' features. This problem is solved using multi-instance learning where training is performed with a bag of instances, chunks in this scenario, instead of one single feature vector. Experiments are performed using ensembles of DNN, CNN and Recurrent Neural Network (RNN). Using ensembles of DNNs, 61.6\% UAR on the development, and 57.4\% UAR on the test set is achieved using the audio modality.

In the assessment of psychiatric disorders, each modality provides new information to the system and increases the diversity in terms of symptoms. Apart from the audio and visual modalities, Zhang et al.~\cite{zhang2020multimodal} propose textual features for mania classification on the BD corpus. For the audio-visual modalities, a Multimodal Deep Denoising Autoencoder (multi-DDAE) framework is proposed to denoise the input and learn shared representations of baseline features (MFCC, eGEMAPS, facial landmarks, eye gaze, head pose, and facial action units). Using fisher vector encoding, the extracted clean feature vectors are encoded into fixed-length vectors. Feature selection using random forest is applied to session-level representation vectors to reduce redundancy and avoid overfitting. For the textual modality, session-level representations are obtained Paragraph Vector and doc2vec models. Early fusion is performed on audio-visual and textual representation vectors. Proposed framework tested on both BD corpus and Extended Distress Analysis Interview Corpus (E-DAIC)~\cite{devault2014simsensei}, which is used in AVEC 2019 challenge~\cite{ringeval2019avec}. Experimental results show that multi-modal frameworks increase the classification performance of mental disorder recognition tasks. On the BD corpus, 70.9\% UAR score is achieved on the development set. Using 10 fold cross-validation, 60.0\% UAR is achieved, which shows that the proposed framework does not overfit to the data.

In ~\cite{sunbipolar} 93.2\% UAR score is achieved using acoustic, visual, and textual modalities on the development set. However, the test set score is not presented. Since the BD dataset is a small one, and prone to overfitting, the test scores should have been submitted to evaluate the system more accurately. For the acoustic features, they use baseline MFCC, eGEMAPS, BoAW, and DeepSpectrum features as well as newly presented Soundnet, which is a one-dimensional fully convolutional network, and ComParE feature set, which is extracted using the openSmile toolkit. For the visual modality, baseline features FAU, BoVW features are used, and CNN, Convolutional Auto-Encoder (CAE), geometric, Multi-scale Dense Scale-invariant feature transform (SIFT), and Histograms of Oriented Gradients-Local Binary Pattern (HOG-LBP) features are presented. Finally, for the textual modality, they use the spaCy toolkit to extract 300 features for each word. Decision level fusion on all modalities is applied after getting the unimodal decision probabilities using SVM classification algorithm. 

\section{Related Works on Weighted Extreme Learning Machine}
\label{section:welm}

In our experiments, we use ELM method as a classification algorithm. BD corpus is an imbalanced dataset, so we experiment Weighted ELM (WELM) method. WELM assigns weights to each sample in a way that it strenghtens the minority class (explained in Section \ref{subsection:methodology} in detail). However, the weights are assigned based on sample quantities, and may not be optimal. 

Wang et.al.\cite{wang2018deep} proposes the Deep WELM (DWELM) method to solve this problem. DWELM consists of enhanced ELM and AdaBoost algorithms. Enhanced ELM is created by replacing the linear ELM with regularized ELM, and adding shortcut connections between the building blocks. An enhanced AdaBoost model embedded into enhanced DWELM algorithm. The AdaBoost algorithm is enhanced in a way that the weights are updated for both the misclassified and correctly classified samples. Their experimental results show that proposed algorithm is efficient on both binary and multiclass classification problems.

In \cite{yao2018weight}, the imbalance learning problem on ELM model is solved using genetic algorithms. They propose a weighted and cost sensitive ELM model. They use the cost matrix in weighted least square method, and assign different weights to each sample. Genetic algorithm is used to obtain the optimal cost. Their experiments show that cost sensitive WLS approach performs better than the WELM model.

\chapter{THE TURKISH AUDIO-VISUAL BIPOLAR DISORDER CORPUS}
\label{chapter:dataset}

In this work, we use the Turkish Audio-Visual Bipolar Disorder (BD) Corpus~\cite{cciftcci2018turkish},
which was also used for the 2018 AVEC Bipolar Disorder and Cross-cultural Affect Recognition Competition~\cite{ringeval2018avec}, as discussed in the previous chapter. Participants were encouraged to achieve the highest performance, considering the baseline performance given by the organizers. 

The BD corpus contains video clips of 46 bipolar disorder patients and 49 healthy controls from the mental health service of a hospital. Mood of the patients evaluated using YMRS and Montgomery-Asberg Depression Rating Scale (MADRS) during 0th, 3rd, 7th and 28th days of the hospitalization and after discharge on the 3rd month. In those days, psychiatrists performed an interview with the patients asking the same questions each time and took audiovisual recordings of the sessions. Annotation was done based on YMRS score~\cite{young1978rating}. YMRS is a continuous clinical interview assessment scale used for rating the severity of manic episodes of a patient. Scores range from 0 to 60 where higher scores represent severe mania. In the BD corpus, bipolar patients are grouped into three classes based on their YMRS score in a session. Grouping is done considering following the scheme where $Y_t$ represents the YMRS score of session t:

\begin{center}
\begin{varwidth}{\textwidth}
\begin{enumerate}
    \item Remission: $Y_t \leqslant 7$
    \item Hypomania:     $7 \textless Y_t \textless 20$
    \item Mania: $Y_t \geqslant 20$
\end{enumerate}    
\end{varwidth}
\end{center}

\begin{figure}[h]%
    \centering
    \includegraphics[width=10cm]{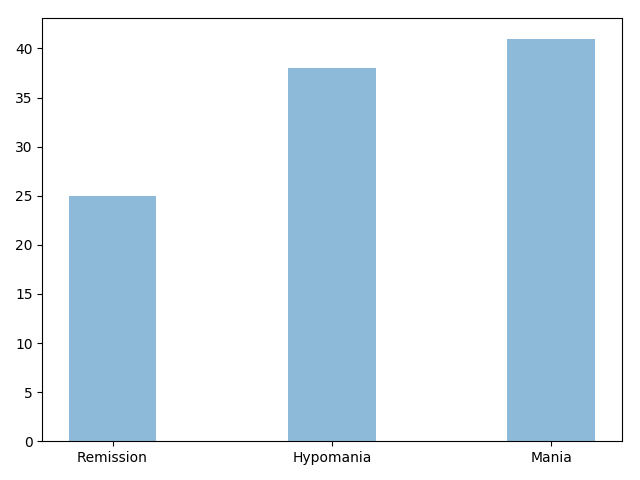}
    \caption{Number of clips per class in the training set}%
    \label{fig:dataset}%
\end{figure}

As presented by the AVEC Competition, there are 104, 60, and 54 clips in the training, development, and test sets, respectively. Due to the difficulties and ethical issues of collecting healthcare data, they are typically small in the number of recordings. So its size should be considered while working on the problem to avoid overfitting and achieve better generalizability. Table~\ref{fig:dataset} shows the distribution of classes in the training set. There are 25, 38, and 41 clips for remission, hypomania, and mania in the training set, respectively. There is a data imbalance in the dataset. It should be handled in order to prevent bias in favor of the majority class.

\begin{figure}[h]%
    \centering
    \subfloat{\includegraphics[width=4cm]{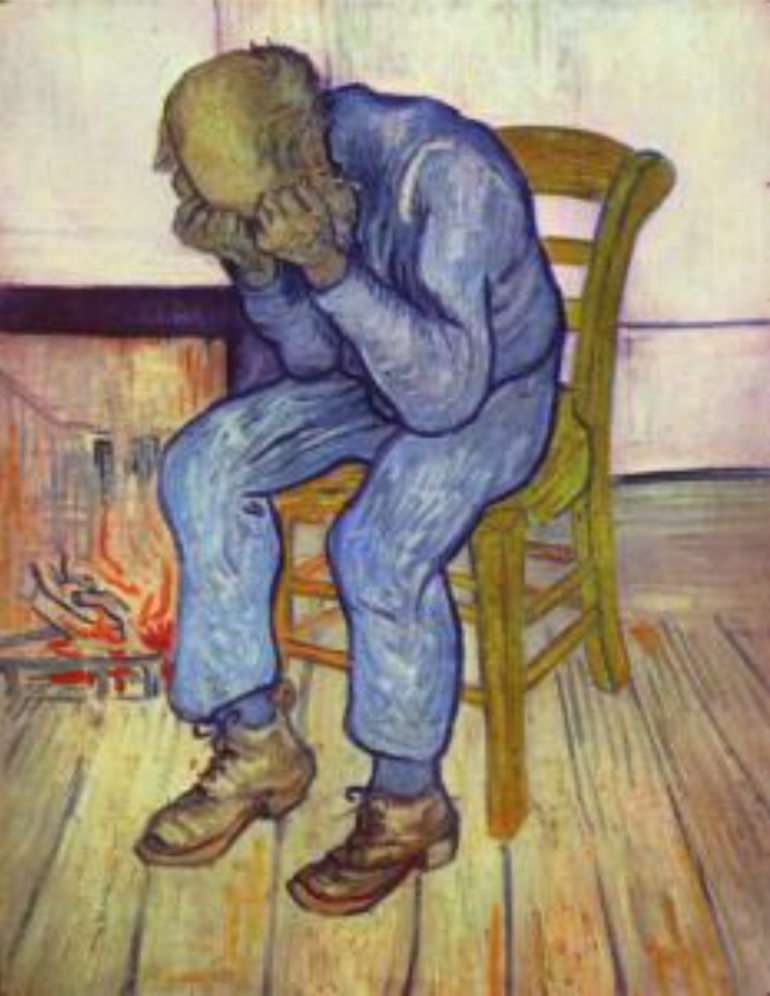} }%
    \subfloat{\includegraphics[width=7cm]{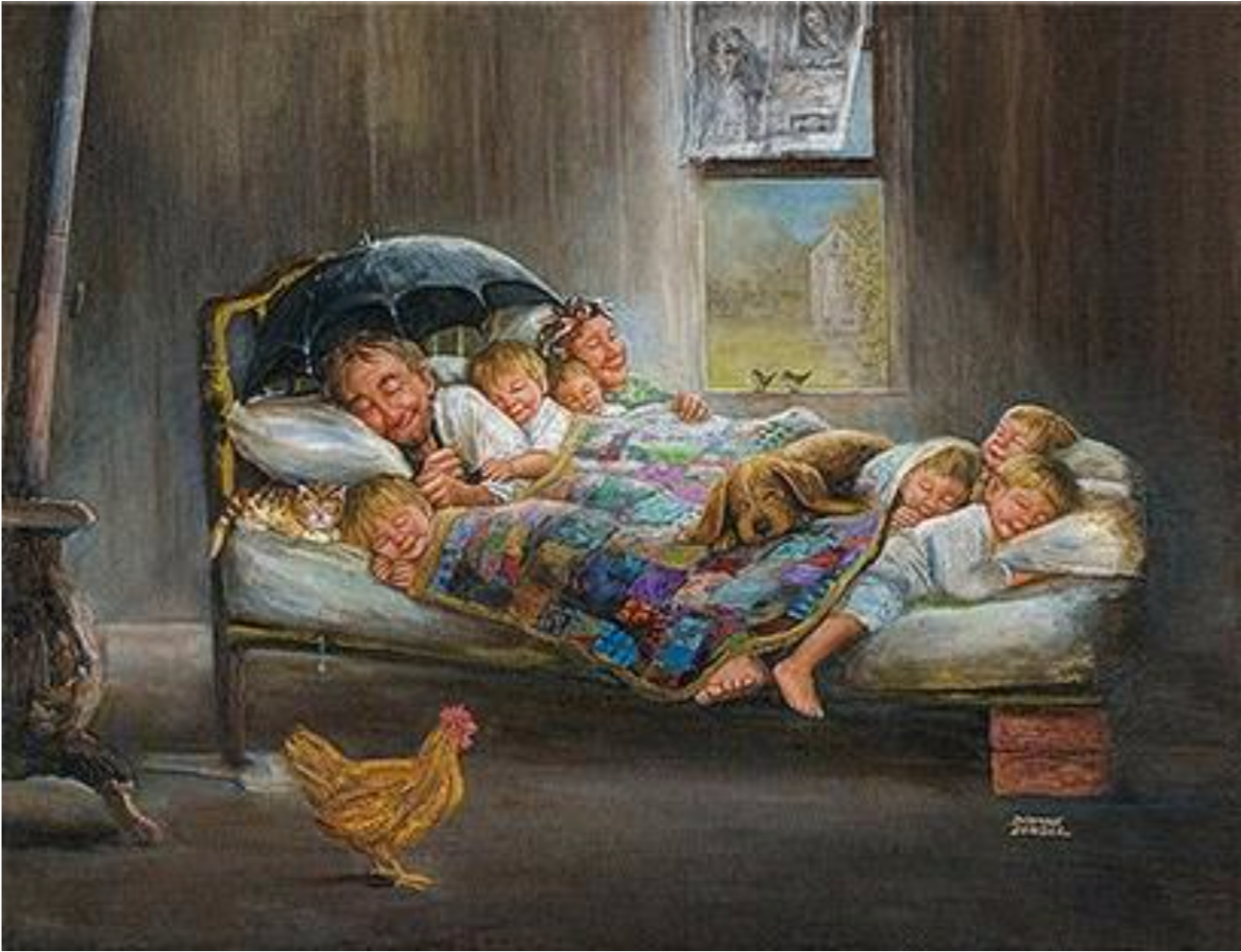} }%
    \caption{Van Gogh's Depression (left), Dengel's Home Sweet Home (right)}%
    \label{fig:pictures}%
\end{figure}

During recordings, patients were asked to perform seven tasks. The tasks were designed to reveal different emotions in the patients so that patients can be observed in different conditions. The first three tasks can be considered as negative emotion eliciting tasks, following two tasks are neutral ones and last two tasks are positive emotion eliciting tasks. The performed tasks are explaining the reason to come to hospital, explaining Van Gogh's Depression picture (see Fig.~\ref{fig:pictures} left), describing a sad memory, counting one to thirty, counting one to thirty faster, explaining Dengel's Home Sweet Home picture (see Fig.~\ref{fig:pictures} right) and  describing a happy memory.

Clips were recorded in a room where only the participant and clinician were present. The participants were recorded with a camera while performing tasks. They read the descriptions of the tasks they were asked to perform from the computer screen. After completing a task, they pushed the space button and a description of the next task appeared on the screen. When the space button was pushed, a `knock' sound was heard to mark the beginning of a new task. This sound helps to split tasks if the tasks are wanted to use separately for classification.

In order to provide baseline results on the data, creators of the corpus investigate audio and visual modalities, experiment with both classification and regression models, and propose two approaches as they call direct and indirect approaches. 

A standard set of audio features are extracted using open-source openSMILE toolkit~\cite{eyben2010opensmile}. openSMILE is a feature extraction tool that is used for extracting large audio feature spaces. For common tasks, it provides example feature sets varying from MFCC for speech recognition tasks to baseline acoustic feature sets of the INTERSPEECH challenges on affect and paralinguistics.

For the BD dataset, 76 dimensional (38 raw, 38 temporal derivative) INTERSPEECH 2010 (IS10) paralinguistic challenge baseline features are used. IS10 configuration file gives a supra-segmental set of 1,582 features, which are calculated using 21 functionals (see a list of functionals on Table~\ref{table:IS10}) on the descriptors (some functionals are not applied to all descriptors). Apart from these suprasegmental features, 10 functionals (BD10 functionals on Table~\ref{table:IS10}) are proposed and applied on 76-dimensional IS10 LLDs, which creates 760-dimensional supra-segmental features~\cite{cciftcci2018turkish}. 

\begin{table}[h]
\begin{center}
\caption[Acoustic features sets and corresponding functionals]{Baseline acoustic feature set provided in IS10 challenge that extracts supra-segmental features using 38 LLDs and 21 functionals listed below. DDP: difference of difference of periods; LSP: line spectral pairs; Q/A: quadratic/absolute.}
\label{table:IS10}
\begin{tabular}{|l|l|l|}
\hline
\multicolumn{1}{|c|}{\textbf{IS10 Descriptors}} & \multicolumn{1}{|c|}{\textbf{IS10 Functionals}} & 
\multicolumn{1}{|c|}{\textbf{BD10 Functionals}} \\ \hline
PCM Loudness                               & Arithmetic mean, Standart dev.   & Mean       \\
MFCC {[}0-14{]}                            & Linear regression coefficients 1/2     & Standard dev.      \\
log Mel Freq. Band {[}0-7{]}               & Linear regression error (Q/A) & Curvature coeff. \\
LSP Frequency {[}0-7{]}                    & Percentile 1/99                   &    Slope and offset      \\
F0 by Sub-Harmonic Sum                     & Percentile range 99-1             &  Min. + relative pos.         \\
F0 Envelope                                & Quartile 1/2/3                    &   Max. + relative pos.        \\
Voicing Probability                        & Quartile range 2-1/3-2/3-1        &    Range (max-min)       \\
Jitter Local                               & Relative pos. min/max             &           \\
Jitter DDP                                 & Skewness, Kurtosis                 &          \\
Shimmer Local                              & Up-level time 75/90                &          \\ \hline
\end{tabular}
\end{center}
\end{table}

For the indirect approach, an emotion classifier is trained on another Turkish emotion corpus and used as a feature extractor. Emotion prediction requires knowledge on varying affective states, so this knowledge can be transferred to the recognition of other mental states both in audio and visual modalities. Bogazici University Emotional Database (BUEMODB) corpus is used for training the emotion classifier~\cite{meral2003analysis}. It contains 484 utterances on four emotions (anger, happiness, neutral, sadness). Apart from the original classifier on four emotion classes, valence and arousal classifiers are also trained. Arousal and valence labels are obtained by labeling the emotion to its relative valence or arousal class. Scores and labels obtained from these emotion classifiers are used as mid-level features for the classification of clips into BD episodes.

Visual features contain both geometric features (GEO) extracted from facial landmarks and appearance features obtained from faces using a pre-trained deep convolutional neural network (DCNN). In order to extract geometric features from faces, 2.2 million frames are collected from clips. Faces in these clips are detected, cropped, registered, and saved as 128x128 gray-scale images. From each face, 23 geometric features are extracted using the method from ~\cite{kaya2015contrasting}. Secondly, appearance features are extracted using a DCNN fine-tuned on an emotion corpus. It gives a 4,096-dimensional feature vector from the last convolutional layer.

\chapter{METHODOLOGY}
\label{chapter:methodology}
In this chapter, we introduce the features used in audio, textual, and visual modalities, preprocessing, feature selection methods applied to the dataset. After that, we explain the ELM algorithm used as a classification method, cross-validation technique used to evaluate the results, and modality fusion methods applied to improve the unimodal results.
\section{Audio Features}
%AAS: bu bölümde suprasegmental, paralinguistic gibi teknik terimleri tanımlayalım. Heysem'in çalışmalarından doğrudan metin kullanmamaya çalışalım, onun stili son derece yoğundur. "Annem okusa anlar mı" diye yazmak en doğrusu. Tane tane, kavramları açıklayarak, mantıksal izlek oluşturarak, argümanları uç uca dizerek yazalım... Çok fazla teknik jargonu cümlelere doldurmayalım...

Feature extraction is the initial stage in most of the machine learning problems where the aim is to obtain representations from the input that can be useful for a pattern recognition process in the further steps. For audio feature extraction, we use the openSMILE feature extraction toolkit~\cite{eyben2010opensmile}, which provides many built-in configuration files that extract the baseline audio features from INTERSPEECH, AVEC challenges, and some parameter sets (GEMAPS) proposed for voice research and affective computing studies on audio.

openSMILE provides a command-line feature extractor program, which takes a configuration file, an input audio file name, an output file name, and some options regarding input and output formats. It accepts audio files in the WAV format. Output can be one of the WEKA Arff, HTK binary, or CSV text formats. Features from configuration files can be extracted as LLDs that gives features for each frame based on window size and length or as supra-segmental features, which are the summaries of LLDs calculated using functionals given on the configuration files. An example command for feature extraction is as follows:

\begin{verbatim}
SMILExtract -C IS10_paraling.conf -I dev_001.wav -csvoutput dev_001.csv
\end{verbatim}
\newpage
Here, \verb SMILExtract \verb   is the command line executable, \verb IS10_paraling.conf \verb   is the path of the configuration file, \verb dev_001.wav \verb   is the path of the input file and \verb dev_001.csv \verb   is the path of the output file. \verb -csvoutput \verb   option extracts summaries of the features and \verb -lldcsvoutput \verb  extracts same features for each time frame.

In our experiments, we use IS10~\cite{schuller2010interspeech}, eGEMAPS~\cite{eyben2015geneva} and MFCC feature sets. IS10 paralinguistic challenge consists of three sub-challenges, namely age, gender, and affect. IS10 feature set was provided to the participants to be used in the audio classification of the sub-challenge problems. It contains 38 low-level descriptors and their temporal derivatives as can be seen in Table~\ref{table:IS10}. 
%pinar, table 3.1 ve 4.1 net oldu mu? burada bir önceki chapter'daki tabloyu soyluyorsun, bir de eGEMAPS tablosu 4.1 var, o da bu paragraftan sonra geliyor. dogru değil mi? ~ işareti unbreakable space demek Latex'de... Table 4.1 yazarken Table ve 4.1'i farklı satırlara yazmasın diye...
%evet buraya da IS10 functionallarının olduğu bir tablo koymuştum ama 3.1'le aynı olduğu için burdakini kaldırdım
The features and their functionals are selected for capturing information relevant to the paralinguistic activity. eGEMAPS is presented as a minimalistic set of audio features compared to large brute-force parameter set (see Table~\ref{table:egemaps_list}). The features are chosen for their ability to represent affective physiological changes in voice production. MFCC features are widely used in speech recognition tasks. They represent the phonemes as the shape of the vocal tract and give information about the human voice perception mechanism. All three feature set LLDs summarised using BD10 functionals used during the experiments were listed in Table~\ref{table:IS10}.

\section{Textual Features}
\label{section:textual_features}
In the recognition of the bipolar disorder, clinicians assess the presence of risk of suicide, risk of violence to persons or property, risk-taking behavior, sexually inappropriate behavior, substance abuse, patient's ability to care for himself/herself, etc.~\cite{hilty1999review}. These can be deducted from what patients say during the interviews of the BD dataset. 
For the textual feature extraction, the text version of the interviews is obtained from audio files using the Google Automatic Speech Recognition (ASR) tool~\footnote{https://cloud.google.com/speech-to-text}. Since the audio files were clipped into tasks for the audio experiments, the transcripts for the tasks were extracted as well. The extracted transcripts contained mistakes since there were words not heard well. So, we manually transcribed the third task, which was describing a sad memory, to further examine the results in a situation where there are no mistakes in the texts.

\begin{table}[H]
\begin{center}
\caption[23 LLDs for eGEMAPS feature set.]{23 LLDs for eGEMAPS feature set. HNR: Harmonics to Noise Ratio.}
\label{table:egemaps_list}
\begin{tabular}{l}
\hline
\textbf{3 energy/amplitude related LLDs} \\ \hline
Loudness                                 \\
HNR                                      \\
Shimmer                                  \\ \hline
\textbf{14 spectral LLDs}                \\ \hline
Alpha ratio (50–1000 Hz/1–5 kHz)         \\
Hammarberg index                         \\
MFCCs 1-4                                \\
Formants 1, 2, 3 (rel. energy)           \\
Harmonic difference H1-H2, H1-A3          \\
Spectral flux                             \\
Spectral slope (0–500 Hz, 0–1 kHz)       \\ \hline
\textbf{6 frequency related LLDs}        \\ \hline
F0 (linear and semi tone)                \\
Jitter (local)                           \\
Formant 1 (bandwidth)                    \\
Formants 1, 2, 3 (frequency)             \\ \hline
\end{tabular}
\end{center}
\end{table}

Transformer language embeddings (GPT-2~\cite{radford2019language}, BERT~\cite{devlin2018bert}, GPT-3~\cite{brown2020language}) are the state-of-the-art natural language processing (NLP) models in representing language features. However, these complex models show unreliable results on small datasets. So, we use three alternative feature sets for the linguistic experiments, which are linguistic inquiry and word count (LIWC)~\cite{pennebaker2001linguistic}, term frequency-inverse document frequency (tf-idf), and polarity features. 

LIWC is a text analysis tool that calculates the linguistic or psychological categories of words where the categories indicate social, cognitive, and affective processes. It was first created in 1993 and updated in 2001, 2007, and 2015 with an expanded dictionary. LIWC2015 version extracts 93 features for an input text file. Before using LIWC, the Turkish transcripts extracted from the patient clips translated into English via Google Translation engine~\footnote{https://cloud.google.com/translate}. LIWC features include word count, four summary language variables (analytical thinking, clout, authenticity, and emotional tone), three general descriptor categories (words per sentence, percent of target words captured by the dictionary, and percent of words in the text that are longer than six letters), 21 standard linguistic dimensions (e.g., percentage of words in the text that are pronouns, articles, auxiliary verbs, etc.), 41 word categories tapping psychological constructs (e.g., affect, cognition, biological processes, drives), six personal concern categories (e.g., work, home, leisure activities), five informal language markers (assents, fillers, swear words, netspeak), and 12 punctuation categories (periods, commas, etc).

Tf-idf is a statistical measure that shows how much a word is important in a document. They are used commonly in NLP~\cite{trstenjak2014knn,van2010automatic}, information retrieval~\cite{hiemstra2000probabilistic} and text mining~\cite{jing2002improved} tasks. As a preprocessing step, stop words are removed using the English/German stop-word dictionaries from the NLTK library~\cite{bird2009natural}, and stemming is applied using the Porter stemmer algorithm~\cite{porter1980algorithm}. After these steps, Tf-idf features are computed over the set of uni-grams and bi-grams.

As polarity features, we use the outputs of three sentiment analysis tools together, which are Natural Language Toolkit Valence Aware Dictionary for sEntiment Reasoning (NLTK Vader)~\cite{gilbert2014vader}, TextBlob~\cite{loria2018textblob} and Flair~\cite{akbik2019flair} since they all have different strengths. NLTK Vader is one of the most popular sentiment analysis tools. It uses sentiment lexicon together with grammatical rules for expressing polarity. A sentiment lexicon is a dictionary, which holds the sentiment scores for words, phrases, and emoticons. However, this approach causes the algorithm to perform weakly on unseen words. The algorithm also handles other linguistic usages that can represent sentiment like capitalization, punctuation, adverbs, etc. using some heuristics.  TextBlob library performs many NLP tasks like tokenization, lemmatization, part-of-speech tagging, finding n-grams as well as sentiment analysis. It returns the sentiment with polarity and subjectivity scores where subjectivity represents the amount of personal and factual information in the sentence, which is a good feature for the valence dimension. However, TextBlob does not consider negation in the sentence for the polarity score, which can be misleading. Flair uses a character-level LSTM network for sentiment analysis, so it is good at assessing the unseen words as well.

Sentiment and subjectivity features obtained from each library are combined into a feature vector. Then each feature is summarized with five functions, namely mean, standard deviation, maximum, minimum, and summation, respectively.

\section{Visual Features}
\label{section:visual_features}
Clinicians gain significant insight from visual cues in the recognition of the bipolar disorder. Some of the scoring items of YMRS can be obtained from visual cues like increased motor activity-energy, irritability, elevated mood, appearance, disruptive-aggressive behavior. Besides, the speech rate and the amount can be also observed in the facial actions. 

For the visual experiments, we use FAUs, geometric features extracted from each face, and appearance descriptors. All three of them were presented by dataset owners as a baseline feature. FAU features were presented in AVEC challenge paper~\cite{ringeval2018avec} and the other two in Asian Conference on Affective Computing and Intelligent Interaction (ACII) paper~\cite{cciftcci2018turkish}. The Facial Action Coding System (FACS) is a way to describe emotion via the movements of specific facial muscles. Each FAU represents a movement of an individual muscle. For example, 1 is inner brow raiser, 15 is lip corner depressor and 27 is mouth stretch. Emotional expressions typically correspond to combinations of various action units. In ~\cite{ringeval2018avec}, intensities of 16 FAUs along with a confidence score are extracted using the OpenFace toolkit~\cite{baltruvsaitis2016openface}. The 16 FAUs extracted for this task are as follows: inner brow raiser (AU1), outer brow raiser (AU2), brow lowerer (AU4), upper lid raiser (AU5), cheek raiser (AU6), lid tightener (AU7), nose wrinkler (AU9), upper lip raiser (AU10), lip corner puller (AU12), dimpler (AU14), lip corner depressor (AU15), chin raiser (AU17), lip stretcher (AU20), lip tightener (AU23), lips parts (AU25) and jaw drop (AU26).

Geometric features for video-based emotion recognition in uncontrolled conditions was first suggested by ~\cite{kaya2015contrasting}. They represented different aspects like distance, angle, or ratio of face landmarks. In ~\cite{cciftcci2018turkish}, 23 geometric features are extracted from faces collected and cropped from the videos. The 23 features are eye aspect ratio, mouth aspect ratio, upper lip angles, nose tip - mouth corner angles, lower lip angles, eyebrow slope, lower eye angles, mouth corner - mouth bottom angles, upper mouth angles, curvature of lower-outer lips, curvature of lower-inner lips, bottom lip curvature, mouth opening/mouth width, mouth up/low, eye - middle eyebrow distance, eye - inner eyebrow distance, inner eye - eyebrow center, inner eye - mouth top distance, mouth width, mouth height, upper mouth height, lower mouth height.

Lastly, in ~\cite{cciftcci2018turkish}, the authors have extracted appearance descriptors from faces using a pre-trained DCNN network trained on a face emotion corpus. As stated in the paper, this approach is applied to emotion and apparent personality trait recognition tasks in uncontrolled conditions and gives promising results~\cite{kaya2017video}. From the last convolutional layer of the DCNN network, 4,096-dimensional features are extracted and then summarised using mean and standard deviation functionals.

\section{Preprocessing}
\label{section:preprocessing}
The feature vectors extracted for each clip contain representations of auditory, visual, and textual signals extracted in various ways. All the features have different ranges and scales. However, the model should not consider the larger numeric values as more important in the decision process. So feature standardization or normalization needs to be performed before model training. Normalization generally means dividing the feature vector by its length. It brings all the values between 0 and 1, thus into a common scale. Standardization means bringing the feature vector into the standard normal distribution by subtracting the mean and dividing it to the standard deviation of the feature vector. 

The features we used for the classification of the clips are represented as two-dimensional matrices where columns are the functionals of the features and each row contains the feature vector of a clip. We experiment with both row level and column level normalizations. On the column level, standardization brings each feature to the same scale, which ensures the comparability of the features. Similarly, on the row level, each clip becomes comparable with normalization.

\section{Feature Selection}
\label{section:feature_selection}

Using a large number of features in classification problems usually helps to increase model performance. Gathering as much information as possible about the data results in a better distinction between classes. However, some features may not be relevant to the problem when using a small dataset and decrease generalization capability on the test set. Using many features may lower training speed, thus increase training time especially on large datasets. Another issue is model explainability.  The explainability of a machine learning model is the ability to explain the logic behind the predictions rather than perceiving the model as a black-box machine. Using all the available features without knowing their importance for the problem decreases the explainability of the model.

Since we use high-dimensional common feature sets considering the sample size of the BD dataset, we experiment with some feature selection methods to prevent overfitting and eliminate the irrelevant features. 

Feature selection is the process of selecting the subset of the features used in training the model. It can help to create a more generalizable model, prevent overfitting, select and remove irrelevant features, make the model more explainable, and reduce the training time. In our case, reducing the training time was not our purpose as the dataset size was already small and training didn't take too much time.

For the experiments we use L1-based and tree-based feature selection method, and principal component analysis (PCA).

\subsection{L1-Based Feature Selection}

Regularization is the process of adding a penalty to the model coefficients to reduce overfitting. While regularizing the linear models with L1 norm some coefficients may become zero. So these features can be removed from the model. Thus L1-based linear model regularization can be used as a feature selection method. In our experiment setup, we used a linear SVM model with an L1 penalty available in the scikit-learn library. 

\subsection{Tree-based feature selection}
Random forests are ensemble learning methods that consist of many decision trees, which only a random subset of the features given to the model. This makes the model prone to overfitting. Each tree tries to split the dataset into two in a way that similar samples remain in the same set. This is done by finding the optimal separation based on the impurities of the features. Impurity is the measure of optimal condition for a feature. While using the random forest for feature selection, the impurity decrease for each feature is found and the features are ranked based on that measure. We used the ExtraTreeClassifier method and treat the subset of features used by the tree ensemble as selected feature set.

\subsection{Principal Component Analysis}
Principal component analysis (PCA) is an unsupervised dimensionality reduction technique, which finds the projection of data points into a lower-dimensional space. It creates a hierarchical coordinate system in a way that captures the maximum variance in the data. We used PCA mostly with very high dimensional data like TF-IDF features in linguistic experiments and features extracted from the DCNN network for the visual modality to reduce the feature set size before the feature selection experiments. We used the PCA method from the scikit-learn library.

\section{Classification}
\label{subsection:methodology}
Like most of the healthcare datasets, the BD dataset is a small one with 164 data points in total. While working with a small number of observations, it is crucial to pay attention to getting accurate predictions by avoiding overfitting. 

Classifier selection is one of the most crucial steps while working with small datasets. Deep learning models have been used in many problems and improve the state-of-the-art results. However, complex models with many parameters require many iterations to optimize their parameters, and this results in overfitting in small datasets. Using simple models is a better choice.

In our experiments, we mostly use kernel ELM~\cite{huang2011extreme}.  ELM is a simple and robust machine learning model that contains a single hidden layer. Input weights are randomly initialized, so they do not need to be tuned. The weights between the hidden layer and the output layer are calculated by an inverse operation. 

In a single hidden layer ELM,, the hidden layer output matrix is $\textbf{H} \in \mathbb{R}^{N \times h}$, the weight matrix between the hidden layer and the output layer is $\beta \in \mathbb{R}^{h \times 1}$ and the output layer matrix is $\textbf{T} \in \mathbb{R}^{N \times 1}$, where $N$ is the number of training samples and $h$ is the number of hidden layer nodes. The output weight matrix $\beta$ is calculated using least squares solution of $\textbf{H}\beta = \textbf{T}$ as $\beta = \textbf{H}^{\dag}\textbf{T}$. $\textbf{H}^{\dag}$ represents the Moore-Penrose generalized inverse~\cite{rao1972generalized}, which minimizes $L_{2}$ norms of both $\| \textbf{H}\beta - \textbf{T}\|$ and $\|\beta \|$. For increased generalization and robustness, a regularization coefficient C is used. So, the set of weights is calculated as:
\begin{equation}
 \beta = \left(\frac{\mathbf{I}}{C}+\mathbf{K}\right)^{-1}\mathbf{T},
 \end{equation}
where $\textbf{I}$ is an identity matrix, and \textbf{K} is a kernel. We use radial basis function (RBF) calculating kernel \textbf{K}, as suggested in~\cite{gurpinar2016kernel}. 
\newpage
While working with small datasets, class imbalance may mislead the model in favor of the majority class. Using weighted models is one solution to the imbalanced learning problem. In weighted ELM~\cite{zong2013weighted}, we define a $N \times N$ diagonal weight matrix \textbf{W}, where $N$ is the number of samples. Each diagonal element stores the multiplicative inverse of the number of training samples with the corresponding label. Integrating \textbf{W} into the formula, the set of weights calculated as:
\begin{equation}
\beta = \left(\frac{\mathbf{I}}{C}+\mathbf{WK}\right)^{-1}\mathbf{WT}.
\end{equation}
There is a trade-off between weighted and unweighted models, where the former improves UAR,  while the latter improves accuracy. To find the best performing model, we implement a decision level fusion model:
\begin{equation}
\mathbf{P}_{fusion} = \alpha \mathbf{P}_{unweighted} + (1-\alpha)\mathbf{P}_{weighted}, 
\end{equation}
where $\textbf{P}$ is an $N \times t$ matrix that contains the class probabilities of each sample. $\alpha$ is a coefficient between 0 and 1. The best $\alpha$ is chosen according to the UAR score of $\textbf{P}_{fusion}$.

\section{Cross Validation}
Cross-validation is a model validation technique where the model is evaluated on its ability to generalize to independent data. The dataset is sampled into training and development set repeatedly and a model is created and tested for each split. The BD dataset is a small one with 104 training and 60 development samples. So it is important to make sure the created model is not just performing well on the development set samples but also is a general solution to the problem at hand. Besides, it is also possible to train the model with more data by reducing the development set size. Our main goal for using the cross-validation was to decide which models should we try on the test set by using both development set and cross-validation results.
\newpage
In cross-validation, the dataset is split into k groups. So it is also called k-fold cross-validation. On each turn, one of the k subsets is used as a development set, and the remaining subsets concatenated into a training set. A model is trained, optimized on each training set, and evaluated on each development set. The predictions on each development set is saved. Finally, the performance is evaluated using the predictions and ground truth (real labels) of the whole dataset. 

The parameter k should be chosen in a way that after splitting the data, both training and development sets are still able to be a representative of the dataset. In our case, k is chosen as 4 which creates a training set with 123 and a development set with 41 samples.

\section{Modality Fusion}
\label{section:modality-fusion}
All these modalities complement each other while processing the information. In affective computing, the datasets mostly contain biological signals, which come from various sensors. All these signals contain some common information that complements each other, as well as some specific information that can not be observed from the other ones.

On the other hand, psychiatrists observe patient's speech patterns like rate, amount, appearance, gestures, motor activity, and change of ideas, topics during the interviews. All of these signs are used to decide the patient's YMRS score and to diagnose BD episodes.

The BD dataset contains both audio and video recording of the patients. From the speech recordings, we also acquire the text version of the interviews using the Google ASR tool. We experiment with audio, speech, and text modalities separately. From the results of single modal experiments, we observe that each modality has both advantages and disadvantages specific to itself. For instance, audio modality gives a better score overall, while the hypomania (the middle) class is not classified correctly. However, the linguistic experiments generally give fewer scores than audio modality results, while all three moods of BD are classified with similar performance. So, using the best performing models of the single modalities, we perform some fusion experiments.

Modality fusion can be performed at two levels. It can be done before classification by combining the features from the different modalities, which is called early fusion or feature level fusion. The other approach is called late fusion or decision level fusion, where the outputs of the models combined using suitable methods.

First, we consider the late fusion methods, since the feature vectors used for the modalities were already larger compared to the dataset size and further concatenating these features may lead to overfitting to the data. For the late fusion, we experiment with majority voting and weighted sum methods.

The majority voting method takes the probability labels obtained from each model and outputs the mostly seen label for a sample. If all three models output a different label for a clip, the output label of the audio modality is assigned for that clip, since in general, audio modality performed better. The labels are calculated as:
\begin{equation}
\mathbf{L}_{fusion} = mode(\mathbf{L}_{model_1}, \mathbf{L}_{model_2}, \mathbf{L}_{model_3}),
\end{equation}
where $\textbf{L}$ is an $N \times 1$ matrix that contains the labels for each video clips and $N$ is the number of samples. We take the mode at each row separately.

We use the weighted sum method for both the fusion of two and three modalities. For the fusion of two modalities, the probabilities of each class for a clip from each model given as input and the final probabilities for each class are obtained as: 
\begin{equation}
\label{equation:weighted_sum_fusion}
\mathbf{P}_{fusion} = \alpha \mathbf{P}_{model_1} + (1-\alpha)\mathbf{P}_{model_2},
\end{equation}
where $\textbf{P}$ is an $N \times t$ matrix that contains the class probabilities of each sample where $N$ is the number of samples and $t$ is the number of classes. $\alpha$ is a coefficient between 0 and 1. The $\textbf{P}_{fusion}$ is chosen according to the best UAR score obtained from the probabilities calculated with $\alpha$ values between 0 and 1. For the fusion of three modalities we apply a variant of Equation~\ref{equation:weighted_sum_fusion}. While choosing the coefficients to multiply with the probabilities of a model, we draw three values from the Dirichlet distribution for 500 times and find the optimum sample that maximizes the UAR of the final fusion model.
\begin{equation}
\label{equation:weighted_sum_fusion_dirichlet}
\mathbf{P}_{fusion} = \alpha_0 \mathbf{P}_{model_1} + \alpha_1\mathbf{P}_{model_2} + \alpha_2\mathbf{P}_{model_3}.
\end{equation}

In Equation~\ref{equation:weighted_sum_fusion_dirichlet}, the alpha values are the elements of the vector drawn from the Dirichlet distribution. A probability density function of a Dirichlet distribution of order $N \geqslant 2$ with parameters $\alpha_1,...,\alpha_n > 0$ is
\begin{equation}
\label{equation:dirichlet}
\dfrac{1}{B(\alpha)}\prod_{i=1}^{N} x_{i}^{(\alpha_i-1)},
\end{equation}
where $B(\alpha)$ is a normalizing factor given in terms of multivariate beta function, and $x_i \in (0,1)$ and $\sum_{i=1}^{N}x_{i}=1$.

Finally, we also experiment with early fusion (feature level fusion) methods. In this approach, the features from different modalities are combined into a single feature vector before the classification. In our experiments, each feature vector that is obtained after the summarization of LLDs is concatenated before the normalization operation.

While selecting the fusion models to try on the test set, we consider both 4-fold cross-validation result of a model and Multimodal 1 (MM1) metric~\cite{d2015review}. MM1 metric measures the improvement in the final fusion model. It is calculated as:
\begin{equation}
\label{equation:mm1}
MM1 = \dfrac{UAR_{fusion}-max(UAR_1 ,UAR_2, UAR_3)}{max(UAR_1 ,UAR_2, UAR_3)},
\end{equation}
where $UAR_{fusion}$ is the UAR score of the fusion model, $UAR_1$, $UAR_2$, and $UAR_3$ are the UAR scores of the models created using single modalities. While calculating the MM1 score, we use 4-fold cross-validation scores, since it gives more robust results. After getting the test set results for the selected fusion models, we calculate MM1 scores using test set UARs.

\chapter{EXPERIMENTS AND RESULTS}
\label{chapter:experiments-and-results}
In this chapter, we present the experiment results in three sections. First, we discuss the unimodal systems for both clip level and task level data types. In the final section, the results of the fusion experiments on the clip level data are presented.
\section{Clip Level Experiments}

\subsection{Audio Classification}
For the clip level audio classification, features are extracted from the whole audio clip. This enables us to extract more informative features than extracting features from a specific part of the clip (like task level feature extraction) since the length of the clip is longer. However, in this way it is hard to understand which parts are contributing to the separation of classes. 

Three different sets of features are used for the clip level classification, which are eGEMAPS, IS10, and MFCC features. eGEMAPS and IS10 contain various speech features selected for the paralinguistic speech research. IS10 contains 76 features (38 LLD and their temporal derivatives), while eGEMAPS is a more minimalistic feature set with 23 features. The baseline results on the BD dataset are presented using IS10 features, but as the dataset is a small one with 164 clips, we want to examine the results on a smaller set of features. We also extract the eGEMAPS features, which are summarized using the functionals mentioned in \cite{eyben2015geneva}. Throughout the text, we mention the original eGEMAPS features that contain 88 features as eGEMAPS, and the one summarized using 10 functionals is mentioned as eGEMAPS10. eGEMAPS can be directly extracted as a feature vector with 'csvoutput' option instead of 'lldcsvoutput' with openSMILE command line interface.

MFCC feature set is also extracted with an openSMILE configuration file, which computes 13 MFCC (0-12) and appends their 13 delta and 13 acceleration coefficients. In total, it extracts 39 LLDs. As this set contains only one kind of acoustic feature, it allows us to see how the model performs with a basic set of features. Besides, it improves the explainability of the model. eGEMAPS10, IS10 and MFCC feature sets are summarized with the BD10 functionals listed in Table~\ref{table:IS10}.  For each audio clip, the final feature vectors contain 88, 230, 760, and 390 features for eGEMAPS, eGEMAPS10, IS10, and MFCC sets, respectively.  

This section shows the experimental results on these feature sets with the ablation studies on the techniques we used in order to increase the performance.

%Normalization
Table \ref{table:normalization_results} shows the results with and without L\textsubscript{2} normalization. Z normalization is applied to each feature separately. After that, L\textsubscript{2} normalization is applied to the feature vector of each clip. The ranges and units of the features vary, so the model may give more importance to the features with bigger numbers. Applying normalization to the feature vector eliminates this effect. As can be seen in the results, applying L\textsubscript{2} normalization improves the performance for both feature sets.

\begin{table}[]
\caption[UAR scores on acoustic modality with and without L\textsubscript{2} normalization]{UAR scores obtained using fusion of kernel ELM and unweighted kernel ELM models on the development set with and without L\textsubscript{2} normalization. Z represents the feature level Z normalization and L\textsubscript{2} represents the L\textsubscript{2} normalization.}
\label{table:normalization_results}
\centering
\begin{tabular}{@{}llll@{}}
\toprule
\textbf{Features} & \textbf{Dimension} & \textbf{Normalization} & \textbf{Development}    \\ \midrule
MFCC     & 390 & Z+L\textsubscript{2}           & 60.3\% \\
MFCC     & 390 & Z            & 52.3\% \\
eGEMAPS10  & 230 & Z+L\textsubscript{2}           & \textbf{63.7}\% \\
eGEMAPS10  & 230 & Z            & 58.2\% \\
 \bottomrule
\end{tabular}
\end{table}

%Feature selection
The dimensions of the features extracted for the audio modality are high considering the sample size of the BD dataset, which can lead the model to overfit the data. Besides, some features may be irrelevant to the problem as we use common feature sets. These irrelevant features may mislead the model and reduce performance. So we experiment with some feature selection methods to prevent overfitting and eliminate the irrelevant features, as explained in Section~\ref{section:feature_selection}.

As can be seen in Table \ref{table:feature_sel_results}, both L\textsubscript{1} based and tree-based feature selection methods improve the performance equally for MFCC feature set. However, for the eGEMAPS feature set, feature selection methods drop the performance.

\begin{table}[]
\caption[UAR scores on acoustic modality with different feature selection methods]{UAR scores obtained using fusion of kernel ELM and unweighted kernel ELM models on the development set without any feature selection and L\textsubscript{1} and Extra Tree Classifier feature selection methods. (Feat. Sel.: Feature Selection Method)}
\label{table:feature_sel_results}
\centering
\begin{tabular}{@{}llll@{}}
\toprule
\textbf{Features} & \textbf{Dimension} & \textbf{Feat. Sel.} & \textbf{Development}    \\ \midrule
MFCC     & 390 &    None         & 60.3\% \\
MFCC     & 77 & L\textsubscript{1}            & 61.1\% \\
MFCC     & 168 & Tree            & 61.1\% \\
eGEMAPS10     & 230 &     None        & 63.7\% \\
eGEMAPS10     & 63 & L\textsubscript{1}           & 47.0\% \\
eGEMAPS10     & 98 & Tree           & 60.8\% \\
 \bottomrule
\end{tabular}
\end{table}

%Feature sets
Table \ref{table:feature_results} shows the effect of the different sets of features with normalization and without any feature selection. Both the training and development sets are transformed onto the distribution of the training set features. As explained in Section~\ref{subsection:methodology}, the decision level fusion of weighted and unweighted RBF kernel ELMs is used for the classification. In this setup, the best result is achieved on eGEMAPS10 features with 63.7\% UAR. 

\begin{table}[h]
\caption[UAR scores on acoustic modality on the final system]{UAR scores obtained using fusion of kernel ELM and unweighted kernel ELM models on the development set. (KELM: Kernel Extreme Learning Machine, W-KELM: Weighted Kernel Extreme Learning Machine)}
\centering
\label{table:feature_results}
\begin{tabular}{@{}llllll@{}}
\toprule
\textbf{Features} & \textbf{Dimension} & \textbf{KELM(C)} & \textbf{W-KELM(C)} & \textbf{Alpha} & \textbf{Development} \\ \midrule
MFCC              & 390                & 10                     & 100                             & 0.50           & 60.3\%       \\
eGEMAPS10           & 230                & 10                     & 10,000                            & 0.90           & \textbf{63.7}\%       \\
IS10              & 760                & 10                    & 1,000                             & 0.75           & 55.2\%       \\
eGEMAPS              & 88                & 10                    & 10                             & 0.0           & 52.9\%       \\\bottomrule
\end{tabular}
\end{table}

\begin{figure}[h]%
    \centering
    \subfloat{\includegraphics[width=8cm]{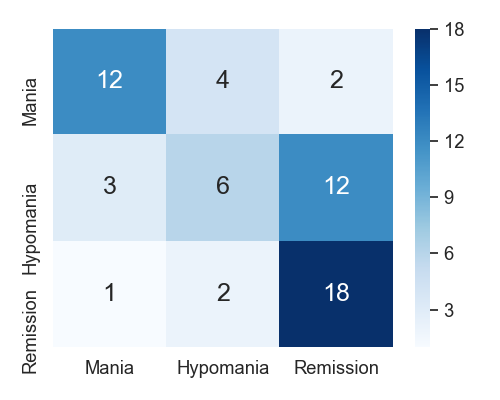} }%
    \subfloat{\includegraphics[width=8cm]{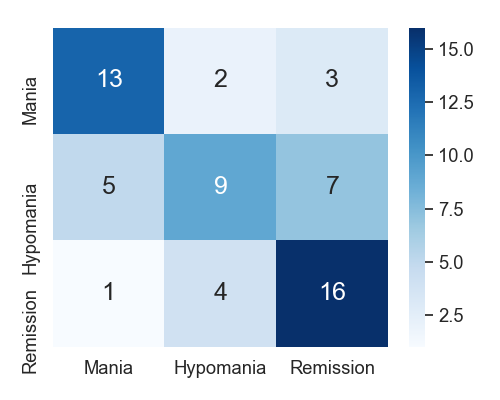} }%
    \caption{Confusion matrices of the results on fusion ELM model with MFCC features (left) and eGEMAPS10 features (right)}%
    \label{fig:mfcc_egemaps_cm}%
\end{figure}

Figure \ref{fig:mfcc_egemaps_cm} shows the confusion matrices obtained from the models created using MFCC and eGEMAPS10 features in Table~\ref{table:feature_results}. The mania and remission classes are classified better than the hypomania class. The YMRS scores between 7 and 20 are labeled as hypomania and it is between the mania and remission classes. So it is harder to differentiate from the other two moods of bipolar disorder. Especially, for the results obtained using the MFCC feature set, the classification performance of the hypomania class is very low. eGEMAPS10 feature set contains various features that are frequency, energy-related, and spectral parameters while MFCC features contain only cepstral parameters. So, a more distinct classification of hypomania class may require more information than cepstral features.

%cross validation
To further examine the behavior of the feature set, we apply k-fold cross-validation to the total of training and development sets. Cross-validation results are not comparable with previous works as the development set changes. However, it is useful while choosing the best model for testing. The k parameter is chosen as 4 in a way that after splitting, both train and holdout sets are large enough to be statistically representative of the data. For the BD dataset, after 4-fold cross-validation, the training set contains 123 samples and the development set contains 41 samples. Results of cross-validation can be seen in Table
~\ref{table:feature_results_cv}. Here, MFCC still shows better performance, while the performance drops for eGEMAPS10. IS10 does not perform well on the development set, but gives the second highest UAR score among other feature set in cross-validation setup, which shows that it can generalize well to the unseen data.

\begin{table}[]
\caption[4-fold cross-validation UAR scores on acoustic modality]{4-fold cross-validation UAR scores obtained using fusion of kernel ELM and unweighted kernel ELM models. (Norm: Normalization, 4F CV: 4-fold cross-validation)}
\label{table:feature_results_cv}
\centering
\begin{tabular}{@{}llll@{}}
\toprule
\textbf{Features} & \textbf{Dimension} & \textbf{Norm.}  & \textbf{4F CV} \\ \midrule
MFCC     & 390  & Z+L\textsubscript{2}  & \textbf{59.2}\% \\
eGEMAPS10  & 230  & Z+L\textsubscript{2}  & 53.1\% \\
IS10     & 760  & Z+L\textsubscript{2}  & 56,8\% \\
eGEMAPS     & 88  & Z+L\textsubscript{2}  & 53.8\% \\\bottomrule
\end{tabular}
\end{table}

\begin{figure}[h]%
    \centering
    \subfloat{\includegraphics[width=8cm]{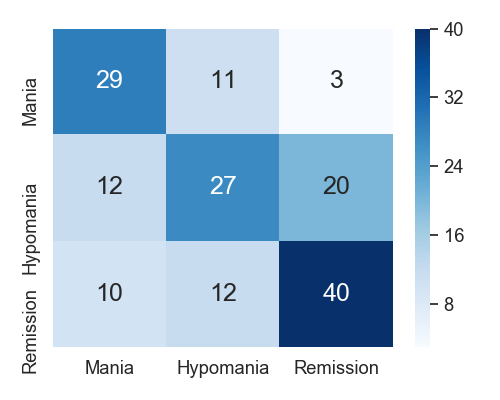} }%
    \subfloat{\includegraphics[width=8cm]{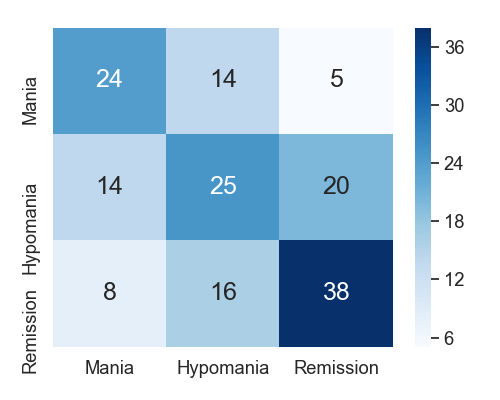} }%
    \caption{Confusion matrices of the results on fusion ELM model with MFCC features(left) and eGEMAPS10 features(right) with cross validation}%
    \label{fig:mfcc_egemaps_cm_cv}%
\end{figure}

Figure \ref{fig:mfcc_egemaps_cm_cv} is the confusion matrix of the results obtained using MFCC and eGEMAPS10 features on the same setup as Figure \ref{fig:mfcc_egemaps_cm} with cross-validation. Since the dataset size changes, it is hard to compare the results with Figure~\ref{fig:mfcc_egemaps_cm}. However, it can be seen that the hypomania class is classified better in this configuration. This indicates that the hypomania class samples in the development set are not distinctive with the audio features. But all three classes can be classified using audio features.

\subsection{Text Classification}
Text level BD classification is performed with the same configuration used in the audio experiments. For the clip level experiments, the text of each clip is obtained from the whole speech file using the Google ASR tool. Since the feature extraction tools used are available for the English language, the texts translated into English using the Google Translation engine. We experiment with LIWC, TF-IDF and polarity features (see Section~\ref{section:textual_features}) for the text classification.

\begin{table}[h]
\caption[UAR scores of text level experiments on the entire clip]{Dev. and 4F CV are the UAR scores obtained on the development sets and on 4-fold cross-validation setup respectively. (Norm: Normalization, Dev: Development, 4F CV: 4-fold cross-validation)}
\label{table:clip_linguistic}
\centering
\begin{tabular}{lllll}
\toprule
\textbf{Features} & \textbf{Dimension} & \textbf{Norm.} & \textbf{Dev.} & \textbf{4F CV}\\ \midrule
LIWC              & 93                 & Z+L\textsubscript{2}                & \textbf{53.7\%} & \textbf{57.3}\%       \\
TF-IDF 500 Bigram & 500                &       None            & 52.9\%  & 48.3\%     \\
Polarity          & 35                 & Z+L\textsubscript{2}                  & 48.9\%     &  42.5\%\\ \bottomrule
\end{tabular}
\end{table}

Table \ref{table:clip_linguistic} shows the results on the text features obtained from the entire clip. All three results are got from the fusion of weighted and unweighted kernel ELM.  

\begin{figure}[h]%
    \centering
    \subfloat{\includegraphics[width=8cm]{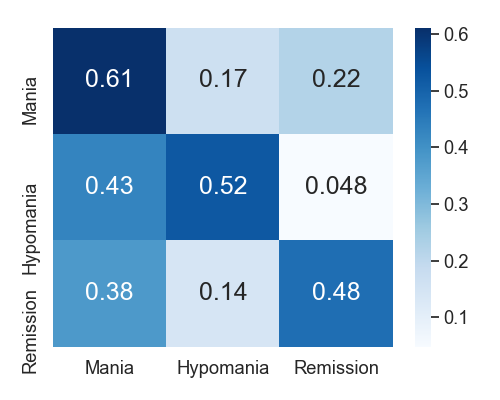} }%
    \subfloat{\includegraphics[width=8cm]{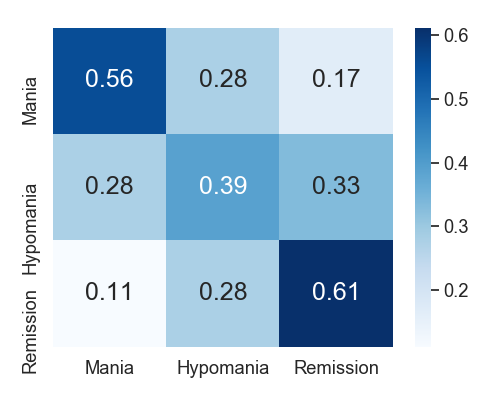} }%
    \caption{Confusion matrices of the results on fusion ELM model with LIWC features on the development set (left) and the cross-validation setup (right).}%
    \label{fig:liwc_cm}%
\end{figure}

In this setup, LIWC features give the best result for both development set and cross-validation experiment with 53.7\% and 57.7\% UAR respectively. Figure~\ref{fig:liwc_cm} shows the confusion matrices obtained using LIWC features on the development set and cross-validation. The increase in the UAR score stems from the higher number of samples on the training set while training the cross-validation model. Though, these results show that LIWC features are successful in the classification of BD episodes and not overfit to the data.

\newpage
\subsection{Video Classification}
To further examine the classification of BD episodes using visual modality and using the results for the fusion of modalities, we experiment with visual features extracted from the video recordings of the patients. For the visual experiments, FAUs, geometric features, and appearance descriptors are used (explained in Section~\ref{section:visual_features}), which were presented by dataset owners as baseline feature sets. We look for the results of summarizing the LLDs of these features using some functionals.

Table \ref{table:clip_visual} shows the best results achieved on the visual modality. All the feature sets are normalized using Z and L\textsubscript{2} normalization as explained in Section~\ref{section:preprocessing}. For the VGG features, 4,096-dimensional features are extracted from the DCNN network, then summarised with mean and standard deviation functionals, which creates an 8,192-dimensional feature vector for each clip. We then reduce the dimensionality using PCA with 99\% variance to 82 dimensions, and apply tree feature selection methods. For the VGG feature set, using only PCA gives the best result. The fourth row in Table~\ref{table:clip_visual} shows the results obtained with a 49-dimensional feature vector after applying PCA to the VGG feature vector. 

\begin{table}[h]
\caption[UAR scores of visual experiments on the entire clip with 4-fold cross-validation]{Dev. and 4F CV are the UAR scores of visual modality experiments obtained on the development sets and on 4-fold cross-validation setup respectively. (Std: Standard Deviation, Dev: Development, 4F CV: 4-fold cross-validation)}
\label{table:clip_visual}
\centering
\begin{tabular}{lllll}
\toprule
\textbf{Features} & \textbf{Functionals} & \textbf{Dimension} & \textbf{Dev.} & \textbf{4F CV}\\ \midrule
GEO & Mean            & 23 & \textbf{57.1\%} & 59.2\% \\
GEO & Mean, Std.  & 46 & 55.8\%          & \textbf{60.7\%}          \\
FAU & Mean, Std.  & 32 & 55.8\%          & 56.0\%        \\
VGG & Mean, Std. & 49 & 41.2\%          & 52.2\%        \\ \bottomrule
\end{tabular}
\end{table}
\newpage
On the development set, the best result is achieved using geometric features summarized using mean functional with 57.1\% UAR on the development set. Using 4-fold cross-validation 60.7\% UAR is achieved on geometric features summarized with mean and standard deviation.

\begin{figure}[h]%
    \centering
    \subfloat{\includegraphics[width=8cm]{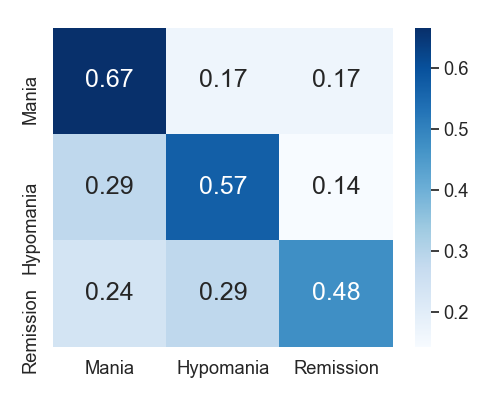} }%
    \subfloat{\includegraphics[width=8cm]{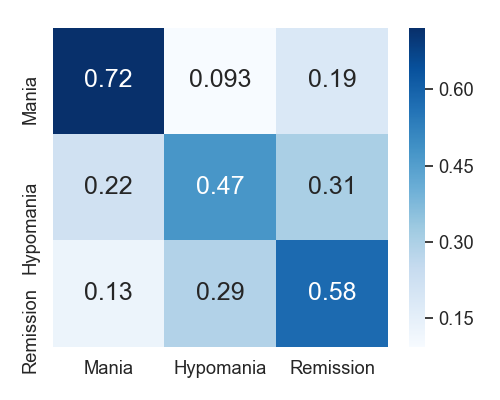} }%
    \caption{Confusion matrices of the results on fusion ELM model with geometric features summarized with mean functional on the development set (left) and the cross-validation setup (right)}%
    \label{fig:geomean_cm}%
\end{figure}

Figure \ref{fig:geomean_cm} shows the confusion matrices obtained from the fusion ELM model trained using geometric features summarized with mean functional. On the cross-validation experiment, mania and remission classes perform better, but hypomania class accuracy drops. However, the overall performance increase, which indicates that the data does not overfit to the geometric features.

\section{Task Level Experiments}

The clips in the BD dataset are recordings of the participants performing seven different tasks, as explained in Chapter~\ref{chapter:dataset}. The tasks were designed to observe participants while thinking about different mindsets. We wanted to examine the effects of these tasks on the classification of BD episodes. 

As mentioned in Chapter~\ref{chapter:dataset}, the tasks are separated with a `knock' sound. However, sometimes participants keep talking about the previous task after the sound was heard or they accidentally push the space button twice, which creates errors in the separation of the tasks. Subsequently, we marked the beginning and end times of the tasks manually. Then, based on these timestamps, new sound files were created. Some participants skipped the tasks with no answer, so not all task files were available for each clip.

Dividing the clips into seven separate tasks shortens the amount of material for learning classifiers per task. However, the number of samples increases, which may increase the generalizability of the model. The trade-off between these two aspects could improve the overall performance.

After creating separate files for each task, eGEMAPS features for acoustic, TF-IDF features for linguistic, and FAU features for visual modality are extracted. Z normalization is applied at the feature level (column-wise) and then L2 normalization is applied along the feature vectors (row-wise). Decision level fusion of unweighted and weighted kernel ELM model is used for the classification. 

First, each task is inspected separately. Training and development sets contain only feature vectors extracted from clips of one task. Since some tasks are not performed in each clip, the number of samples in the development sets become less than the original development set. For each task from 1 to 7, the training sets contain 97, 101, 78, 100, 97, 98, and 88 samples, respectively, which affects the performances of the models created with these training sets. To make the results directly comparable with other works, the missing samples are considered as hypomania class (the middle class).

\begin{table}[h]
\caption[UAR scores obtained from the each task separately using eGEMAPS, TF-IDF and FAU feature sets]{UAR scores obtained from the each task separately on eGEMAPS, TF-IDF and FAU feature set. The tasks are explained in Chapter~\ref{chapter:dataset}.}
\label{table:task_results}
\centering
\begin{tabular}{llll}
\toprule
\textbf{Task} & \textbf{Acoustic} & \textbf{Linguistic} & \textbf{Visual} \\ \midrule
1 & \textbf{52.3\%} & \textbf{44.9\%} & 39.6\% \\
2 & 44.4\% & 42.8\% & 43.3\%           \\
3 & 41.0\% & 40.7\% & 44.7\%         \\
4 & 42.0\% & 40.2\% & 40.2\%  \\
5 & 40.4\% & 37.3\% & 37.3\%  \\
6 & 45.7\% & 42.5\% & \textbf{46.0\%} \\
7 & 44.1\% & 41.5\% & 44.7\% \\
All & 42.0\% &49.4\% & 52.3\% \\ \bottomrule
\end{tabular}
\end{table}

Table~\ref{table:task_results} shows the UAR scores obtained from all three modalities on each task. Task 5, which is counting one to thirty in a fast way, results are the lowest ones on all three modalities. For the audio classification, counting fast eliminates the distinguishing features of speech for different moods. Both counting tasks (4 and 5) give the lowest results for the linguistic modality, since, in most of the clips there are only numbers and they do not contain distinguishing meanings for the text modality. The same result can be seen in the video modality as well. The duration of the tasks is very small and patients only focus on counting, mostly they don't use facial gestures.

Since the task 3 is only performed in 78 clips, the results on acoustic and linguistic experiments do not give good results. However, for the visual modality, it still gives better results. Task 1 on acoustic and linguistic, and task 6 on visual modality experiments give the best results among the other tasks. Results show that the emotion eliciting tasks are more helpful for the classification of BD moods, as explaining happy or sad memories requires various changes in both facial units, speech patterns, and vocabularies. 

The `All' row shows the results where the models are trained on all the tasks, in other words on 662 samples, which increases the training set size. After getting the results on the development set, we take the mean of the probabilities for all the tasks in a clip. So, all the results in Table~\ref{table:task_results} are presented for 60 samples to make them comparable with each other. As the training set size and the information used for the classification increases, the overall performance improves for text and audio modalities. However, the same result is not observed on the audio modality.

Figure  shows the UAR scores of experiments on this setup. Numbers on the x-axis represent the tasks. Counting tasks yield the lowest results, while the remaining emotion eliciting tasks yield higher results. Tasks 2 (describing a sad memory) and 7 (explaining Home Sweet Home picture) give 53.9\% and 49.2\% UAR scores respectively. We use these two tasks together to train a new model. For each clip, we take the average of the class probabilities of 2nd and 7th tasks, again with the goal of making the results comparable with other works. This gives a UAR of 57.1\%, which is higher than the individual scores.

The tasks performed during the recordings fall into three categories as explained in Chapter~\ref{chapter:dataset}. The first three tasks are negative, the last two are positive emotion-inducing tasks while the fourth and fifth are neutral tasks. We perform experiments on tasks grouped according to their emotions. For the task experiments, LLD features were divided into tasks using the manually labeled time stamps. Then, LLD features for tasks 1-2-3, 4-5, and 6,7 concatenated row-wise, where each row contains a feature vector for a frame. If a task is not performed in a clip, that emotion group is created with the remaining tasks. For instance, if the second task is not performed, a negative emotion task group is created by concatenating the first and third tasks. 

\begin{table}[]
\caption[UAR scores of experiments on the tasks grouped based on inducing emotions on the development set]{UAR scores of experiments on the tasks grouped based on inducing emotions on the development set. Negative, neutral and positive represents the concatenation of the tasks 1-2-3, 4-5 and 6-7, respectively.}
\label{table:task_emotion}
\centering
\begin{tabular}{ll}
\toprule
\textbf{Tasks} &  \textbf{Development}    \\ \midrule
Negative       &  47.3\% \\
Neutral        &  41.5\% \\
Positive       &  43.3\% \\
Neg.+Neut.+Pos.&  \textbf{49.4\%} \\ \bottomrule
\end{tabular}
\end{table}

Table \ref{table:task_emotion} shows the experimental results on grouped tasks. First three rows are the experiment results using only one task group. Among the three, negative tasks give the best result with 47.3\% UAR, while neutral tasks give 41.5\%. The negative group contains three tasks, so the clips are longer and contain more information. Similar to single task experiments, the neutral group (counting tasks) does not give much information about the bipolar disorder moods. 

In the fourth row, the emotion-based grouped clips are used together to train an ELM model. The number of clips in the training set becomes 302, which is almost three times more than using the entire clip without any separation or using tasks separately. The mean of class probabilities of the tasks obtained from a clip are averaged. The results are improved using all emotion groups in the training together.

\section{Fusion of Modalities}
After getting the results on single modality experiments, we perform some fusion experiments using weighted sum, majority voting, and feature fusion methods as explained in Section~\ref{section:modality-fusion}. Mostly, we select the features that are performed well in the single modality experiments, then use them in the fusion modality. For the acoustic modality, we select eGEMAPS10 and eGEMAPS feature sets, since the eGEMAPS feature set is created specifically for the paralinguistic tasks, and has better explainability compared to the MFCC feature set. For the linguistic modality LIWC features, and for the visual modality, FAU and geometric features are used for the fusion experiments. Best performing fusion models on the development set are tested on the testing set as well. Test set results are obtained from the models trained on the training and development sets together, which increases the number of training samples.

The previous works on this dataset use validation set to optimize the model. However, their results show that the performances achieved using validation set for the optimization does not correlate with the results obtained from the test set results, as they can not perform better than the baseline test set result. So, we use 4-fold cross-validation and MM1 scores while selecting the models that will be used for the test set submissions.

\begin{table}[]
\caption[The best fusion model results sorted according to 4-fold cross-validation results]{The best fusion model results sorted according to 4-fold cross-validation results. * indicates that tree selection method is applied to that feature set (4F CV: 4-fold cross-validation, MV: Majority Voting, FF: Feature Fusion, WS: Weighted Sum)}
\label{table:fusion}
\makebox[\textwidth]{%
\begin{tabular}{llllll}
\toprule
\textbf{Fusion} &  \textbf{Acoustic}  & \textbf{Linguistic}  & \textbf{Visual} & \textbf{4F CV} & \textbf{MM1 (4F CV)}\\ \midrule
MV & eGEMAPS10*       & LIWC  & FAU  & \textbf{65.7\%} & \textbf{0.15} \\
MV & eGEMAPS  & LIWC  & FAU  & 65.1\%          & 0.14 \\
MV & eGEMAPS10        & LIWC  & FAU  & 64.8\%          & 0.13 \\
MV & eGEMAPS  & LIWC  & GEO  & 64.7\%          & 0.09 \\
MV & eGEMAPS* & LIWC  & GEO  & 64.4\%          & 0.09 \\
MV & eGEMAPS10        & LIWC  & GEO  & 63.3\%          & 0.07 \\
FF & eGEMAPS* & LIWC  & GEO* & 62.8\%          & 0.06 \\
FF & eGEMAPS  & LIWC  & GEO  & 62.4\%          & 0.05 \\
MV & eGEMAPS* & LIWC* & FAU  & 62.2\%          & 0.10 \\
FF & eGEMAPS* & LIWC  & FAU  & 62.0\%          & 0.08 \\
FF & eGEMAPS* & LIWC  & GEO  & 61.3\%          & 0.04 \\
FF & eGEMAPS  & LIWC  & FAU  & 60.6\%          & 0.06 \\
WS & eGEMAPS  & LIWC  & FAU  & 60.3\%          & 0.05 \\
FF & eGEMAPS* & LIWC  & None & 60.1\%          & 0.05 \\
MV & eGEMAPS10        & LIWC* & FAU  & 59.7\%          & 0.07 \\\bottomrule
\end{tabular}}
\end{table}

Table~\ref{table:fusion} shows the results obtained from the fusion models that achieve higher than 60\% UAR on the 4-fold cross-validation experiments. The table is sorted according to the 4-fold cross-validation results, but corresponding MM1 scores almost sorted as well. This shows that there is correlation between the success of the final model, and the contribution of multimodality to the unimodal systems. 

The best UAR score on the 4-fold cross-validation is achieved using eGEMAPS10 with tree feature selection, LIWC, and FAU features fused with the majority voting method, where 65.7\% UAR score is achieved. MM1 score shows that fusion of the modalities increase the maximum unimodal performance by 15\%, which is the highest MM1 score achieved on the 4-fold cross-validation results as well.

The first six best performing models uses majority voting method, which shows the effectiveness of this method. Even the remaining two majority voting lines in Table~\ref{table:fusion} has higher MM1 score than almost all of the feature fusion lines, which indicates that majority voting contributes most to the unimodal results. Feature fusion method is not as successful as majority voting in increasing the unimodal results, since after concatenating the feature sets, the newly generated feature vector has higher dimension, which requires more data for a robust training~\cite{gunes2008automatic}.

Final test sets experiments are done using the top performing four multimodal fusion systems (first four lines in Table~\ref{table:fusion}), as we wish to have a maximum of 10 test set probes. In order to calculate their MM1 scores on the test set, we also obtain the test set results of the constituent unimodal models, namely for eGEMAPS10, eGEMAPS10 with tree feature selection, eGEMAPS, LIWC, FAU, and geometric features.

\begin{table}[]
\caption[Test set results of the feature combinations, which give the best 4-fold cross validation results on the fusion experiments]{Test set results of the feature combinations, which give the best 4-fold cross validation results on the fusion experiments. * indicates that tree selection method is applied to that feature set. (MV: Majority Voting, 4F CV: 4-fold cross-validation)}
\makebox[\linewidth]{
\label{table:fusion_test}
\begin{tabular}{llllllll}
\toprule
\textbf{Fusion} &  \textbf{Acoustic}  & \textbf{Linguistic}  & \textbf{Visual} & \textbf{4F CV} & \textbf{MM1} & \textbf{Test} & \textbf{MM1} \\ \midrule
MV & eGEMAPS10*      & LIWC & FAU & \textbf{65.7\%} & 0.15 & 61.1\% & 0.10\\
MV & eGEMAPS & LIWC & FAU & 65.1\%          & 0.14 & 57.4\% & 0.0\\
MV & eGEMAPS10       & LIWC & FAU & 64.8\%          & 0.13 & \textbf{64.8\%} & 0.09\\
MV & eGEMAPS & LIWC & GEO & 64.7\%          & 0.09 & 53.7\% & -0.06\\ \bottomrule
\end{tabular}}
\end{table}

Table \ref{table:fusion_test} shows the test set results obtained from the feature combinations, which give the best 4-fold cross validation results on the fusion experiments. The best test set result in achieved using the eGEMAPS10, LIWC and FAU feature sets with the majority voting method. On this setup we achieve 64.8\% UAR score, which is 7.4\% higher than the best performing result published so far. 

\begin{table}[]
\caption[Unimodal 4-fold cross-validation and test set results of the features that are the constituents of the best performing fusion systems]{Unimodal 4-fold cross-validation and test set results of the features that are the constituents of the best performing fusion systems. * indicates that tree selection method is applied to the corresponding feature set. (4F CV: 4-fold cross-validation)}
\label{table:unimodal_test}
\makebox[\textwidth]{%
\begin{tabular}{llll}
\toprule
\textbf{Modality} &  \textbf{Feature Set}  & \textbf{4F CV}  & \textbf{Test}\\ \midrule
Acoustic   & eGEMAPS10*      & 52.2\% & 55.5\% \\
Acoustic   & eGEMAPS & 53.8\% & 57.4\% \\
Acoustic   & eGEMAPS10       & 53.1\% & 59.2\% \\
Linguistic & LIWC          & 57.3\% & 51.8\% \\
Visual     & FAU           & 52.1\% & 51.8\% \\
Visual     & GEO           & 56.5\% & 51.8\% \\\bottomrule
\end{tabular}}
\end{table}

Table~\ref{table:unimodal_test} shows the test set results of the model obtained using the feature sets that perform the best on the multimodal fusion experiments. The test set results are obtained using the model trained on the combination of the training and the development sets, with the parameters optimized on the 4-fold cross-validation experiments. Since test set results are obtained using more data compared to the cross-validation setting, some test set results give higher UAR score than the 4-fold cross-validation results. eGEMAPS10 gives the highest unimodal UAR score, which is also higher than the state-of-the-art test set score on this dataset. 

\begin{figure}[h]%
    \centering
    \subfloat{\includegraphics[width=8cm]{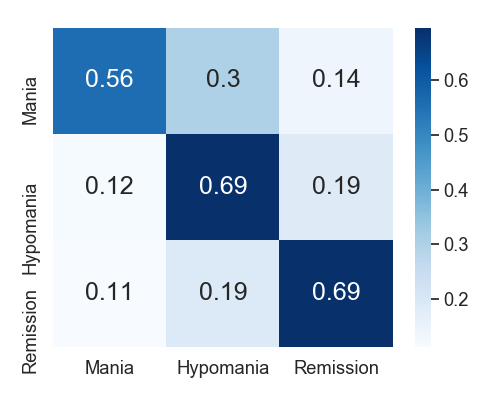} }%
    \subfloat{\includegraphics[width=8cm]{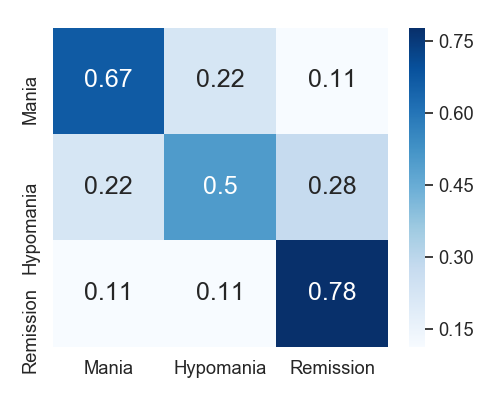} }%
    \caption{Confusion matrices of the best performing result on the test set. The left image is the confusion matrix of the 4-fold cross-validation, and the right image is the confusion matrix of the test set.}%
    \label{fig:best_cm}%
\end{figure}

Figure~\ref{fig:best_cm} shows the confusion matrices on the fusion system that gives the best test set results (LIWC, FAU, and eGEMAPS10 features with majority voting method). We achieve 64.8\% on both 4-fold cross-validation and test set results. From the test set confusion matrix, we can see that mania and remission classes are easier to classify as they have more distinct features. 

\chapter{CONCLUSION}
\label{chapter:discussion}
During this thesis, we worked on the classification of bipolar disorder episodes (mania, hypomania, depression) using the BD dataset that contains video recordings of the bipolar disorder patients while they are interviewed by their psychiatrists. During the interviews, the patients perform seven different tasks. The tasks are designed in a way that they elicit both positive and negative emotions in the patients, and some tasks are emotionally neutral. 

We showed that multimodality improves the generalizability of the classification of bipolar disorder. The information coming from acoustic, textual, and visual modalities complement each other and improve the performance of the unimodal systems. The results suggests that using all three modalities together gives the best performance, however a fusion model of the linguistic, and acoustic modalities still perform well while requiring less information.

%We perform experiments on acoustic, linguistic, and visual modalities separately, and as a multimodal system. 
As a classification algorithm, we use fusion of weighted and unweighted ELMs. ELM was a good fit for this problem, since it is a 2-level neural and prone to overfitting. The data imbalance creates a need for a weighted model, however weighted ELM mostly favor the minority class. So using the fusion of weighted and unweighted ELMs, the optimum point is found. 

\begin{figure}[h]%
    \centering
    \includegraphics[width=11cm]{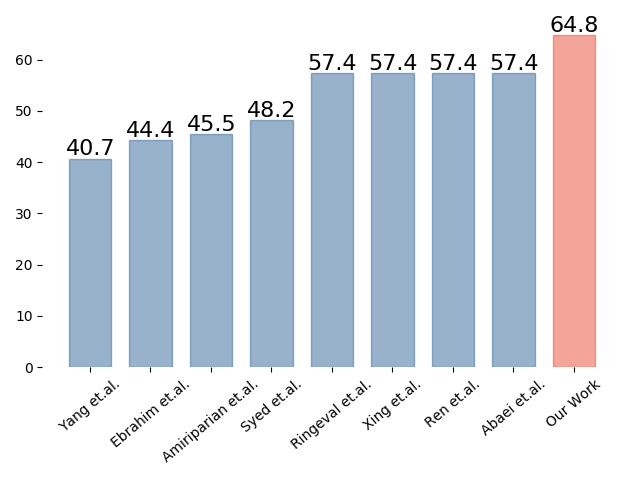}
    \caption{Test set UAR Performance Comparison on BD Dataset}%
    \label{fig:works}%
\end{figure}

The best performing model is achieved using eGEMAPS10, LIWC, and FAU features using the fusion of weighted and unweighted kernel ELMS, and fused using majority voting as a late fusion process. We achieve 64.8\% test set UAR on this configuration, which is the best result achieved on the BD dataset as can be seen in Figure~\ref{fig:works}. The results suggest that benefiting from all three modalities is useful, since the first 13 best performing model is achieved on the fusion models of three modalities. However, the 14th highest score on the Table
~\ref{table:fusion} uses only linguistic and acoustic modalities. So, it is possible to use only audio recordings of the patients, like phone recordings and achieve promising results from the fusion of linguistic and acoustic modalities. Besides, the MM1 scores on Table \ref{table:fusion} shows that, fusion of modalities increase the maximum scores achieved on a single modality in all the configurations.  

eGEMAPS is a commonly used minimalistic acoustic feature set. So we used it for the audio classification, and in the fusion experiments. Besides, we summarized eGEMAPS LLDs with the 10 functionals presented in~\cite{cciftcci2018turkish}. We achieved a better performance using eGEMAPS10 feature set, which shows that eGEMAPS LLDs can give better results when summarized with different functionals. eGEMAPS, and eGEMAPS10 feature sets contain 88, and 230 features respectively. So, a higher feature size may help finding better features that generalize better to the dataset. 

These results are still not high enough to use in a real-world application as a decision system. One of the main difficulties was the small size of the BD corpus. There are 25, 38, and 41 clips in the dataset for the remission, hypomania, and mania classes respectively, which is not enough to generalize with a high certainty. The dataset is collected in a real-life scenario. So there were some noises, and in some cases the clinician explains things about the questions to the patients, so her voice can be heard as well. These issues are expected to be present if a real-life application is created, so the natural recording setup makes this database valuable. Another difficulty stems from missing information in some clips, where patients do not answer some of the questions. In one of the test case clips, the patient does not answer any question at all. This can be used as a feature as well. However in our method, it caused a poor performance.

Besides the clip level evaluation, we look for the effect of the tasks separately, and by grouping the same emotion eliciting tasks during the classification. Since some tasks are not performed in every clip, the number of clips per task are different. To be able to compare the results among the task groups and the entire clips results, we assign the middle class label to the missing clips. Since the dataset size is already small, this distorted the final scores somewhat. Still, from the task level experiments we can see that emotion eliciting tasks are more useful in the classification of BD for all three modalities, as expected. In order to increase the dataset size, we also used the task groups as separate data points and performed classification. However, the results were not better than the entire clip level results, which shows that the information obtained from longer clips is necessary for learning.

Our final best performing model contains information from three different modalities, and each modality is represented using feature vectors with various sizes, which causes poor explainability of the model. It is especially important to create explainable models in medical domain. 
As a further study, the explainability of the system can be investigated, which also gives insights to the psychiatrists about the features used in the classification, and the best performing ones can be adapted in their decision making progresses.

%Bulguların neler söylediğini tartışıyoruz. Hangi yöntem ne kadar iyi çalışmış, işe yarar mı, zorluklar neler, sorunlar neler, neleri iyi yapabildik, nelere daha ihtiyaç var, yeni veri toplanacaksa ne toplanmalı, yeni metot geliştirilecekse ne geliştirilmeli vs.

\begin{comment}
\chapter{CONCLUSIONS}
\label{chapter:conclusions}

In this work, we showed that multimodality improves the generalizability of the classification of bipolar disorder. The information coming from acoustic, textual, and visual modalities are complement each other and improve the performance of the unimodal systems. The results suggests that using all three modalities together gives the best performance, however a fusion model of the linguistic, and acoustic modalities still perform well while requiring less information.

Our final best performing model contains information from three different modalities, and each modality is represented using feature vectors with various sizes, which causes poor explainability of the model. It is especially important to create explainable models in medical domain. 
As a further study, the explainability of the system can be investigated, which also gives insights to the psychiatrists about the features used in the classification, and the best performing ones can be adapted in their decision making progresses.
%tezin katkıları neler oldu, neler öğrendik, bu çalışmanın devamı nereye gider.
\end{comment}

\appendix
\chapter{YOUNG MANIA RATING SCALE (YMRS)}
The list of items used in the scoring of the YMRS scores. Each item is graded a score between 0-4.
\begin{enumerate}[label=\arabic*.]
  \item Elevated Mood
  \item Increased Motor Activity-Energy
  \item Sexual Interest
  \item Sleep
  \item Irritability
  \item Speech (Rate and Amount)
  \item Language-Thought Disorder
  \item Content
  \item Disruptive-Aggressive Behavior
  \item Appearance
  \item Insight
\end{enumerate}

%The appendices start here.
%~\cite{*}
\bibliographystyle{fbe_tez_v11}
\bibliography{references}

\begin{thebibliography}{10}
\newcommand{\enquote}[1]{``#1''}
\expandafter\ifx\csname url\endcsname\relax
  \def\url#1{{\tt #1}}\fi
\expandafter\ifx\csname urlprefix\endcsname\relax\def\urlprefix{}\fi

\bibitem{merikangas2007lifetime}
Merikangas, K.~R., H.~S. Akiskal, J.~Angst, P.~E. Greenberg, R.~M. Hirschfeld,
  M.~Petukhova and R.~C. Kessler, \enquote{Lifetime and 12-month prevalence of
  bipolar spectrum disorder in the National Comorbidity Survey replication},
  {\em Archives of General Psychiatry\/}, Vol.~64, No.~5, pp. 543--552, 2007.

\bibitem{world2008global}
Organization, W.~H. {\em et~al.\/}, {\em The Global Burden of Disease: 2004
  update\/}, World Health Organization, 2008.

\bibitem{lish1994national}
Lish, J.~D., S.~Dime-Meenan, P.~C. Whybrow, R.~A. Price and R.~M. Hirschfeld,
  \enquote{The National Depressive and Manic-depressive Association (DMDA)
  survey of bipolar members}, {\em Journal of Affective Disorders\/}, Vol.~31,
  No.~4, pp. 281--294, 1994.

\bibitem{elvan2017}
Çiftçi, E., {\em Bipolar Bozukluk - Mani Dönemi Tanılı Bireylerin
  Görüntü-Ses Özniteliklerinin Klinink Özellikler ve Nörokognitif
  İşlevlerle İlişkileri\/}, Ph.D. Thesis, Sağlık Bilimleri Üniversitesi
  / İstanbul Erenköy Ruh ve Sinir Hastalıkları Eğitim ve Araştırma
  Hastanesi / Ruh ve Sinir Hastalıkları Ana Bilim Dalı, 2017.

\bibitem{shatte2019machine}
Shatte, A.~B., D.~M. Hutchinson and S.~J. Teague, \enquote{Machine learning in
  mental health: a scoping review of methods and applications}, {\em
  Psychological Medicine\/}, Vol.~49, No.~9, pp. 1426--1448, 2019.

\bibitem{cciftcci2018turkish}
{\c{C}}ift{\c{c}}i, E., H.~Kaya, H.~G{\"u}le{\c{c}} and A.~A. Salah,
  \enquote{The {Turkish} audio-visual bipolar disorder corpus}, {\em 2018 First
  ACII Asia\/}, pp. 1--6, IEEE, 2018.

\bibitem{schimmel2019mp}
Schimmel, A., M.~Doyran, P.~Baki, K.~Ergin, B.~T{\"u}rkmen, A.~A. Salah,
  S.~Bakkes, H.~Kaya, R.~Poppe and A.~A. Salah, \enquote{MP-BGAAD: Multi-Person
  Board Game Affect Analysis Dataset}, {\em 15th International Summer Workshop
  on Multimodal Interfaces\/}, p.~1, 2019.

\bibitem{jmui}
Doyran, M., A.~Schimmel, P.~Baki, K.~Ergin, B.~Türkmen, A.~Akdağ~Salah,
  S.~Bakkes, H.~Kaya, R.~Poppe and A.~A. Salah, \enquote{MUMBAI: Multi-Person,
  Multimodal Board Game Affect and Interaction Analysis Dataset}, {\em Journal
  on Multimodal User Interfaces\/}, 2021.

\bibitem{olalere21hbu}
Olalere, F., M.~Doyran, R.~Poppe and A.~A. Salah, \enquote{Geeks and guests:
  Estimating player's level of experience from board game behaviors}, {\em 11th
  Int. Workshop on Human Behavior Understanding\/}, 2021.

\bibitem{baltruvsaitis2016openface}
Baltru{\v{s}}aitis, T., P.~Robinson and L.-P. Morency, \enquote{Openface: an
  open source facial behavior analysis toolkit}, {\em 2016 IEEE Winter
  Conference on Applications of Computer Vision (WACV)\/}, pp. 1--10, IEEE,
  2016.

\bibitem{baki20}
Baki, P., H.~Kaya, E.~{\c C}ift{\c c}i, H.~G{\" u}le{\c c} and A.~A. Salah,
  \enquote{Speech Analysis for Automatic Mania Assessment in Bipolar Disorder},
  {\em 28th Signal Processing and Communications Applications Conference\/},
  2020.

\bibitem{ojeme2016selecting}
Ojeme, B. and A.~Mbogho, \enquote{Selecting learning algorithms for
  simultaneous identification of depression and comorbid disorders}, {\em
  Procedia Computer Science\/}, Vol.~96, pp. 1294--1303, 2016.

\bibitem{hou2016big}
Hou, Y., J.~Xu, Y.~Huang and X.~Ma, \enquote{A big data application to predict
  depression in the university based on the reading habits}, {\em 2016 3rd
  International Conference on Systems and Informatics (ICSAI)\/}, pp.
  1085--1089, IEEE, 2016.

\bibitem{brasil2009towards}
Brasil~Filho, A.~T., P.~R. Pinheiro and A.~L. Coelho, \enquote{Towards the
  early diagnosis of Alzheimer’s disease via a multicriteria classification
  model}, {\em International Conference on Evolutionary Multi-Criterion
  Optimization\/}, pp. 393--406, Springer, 2009.

\bibitem{labate2014eeg}
Labate, D., F.~La~Foresta, I.~Palamara, G.~Morabito, A.~Bramanti, Z.~Zhang and
  F.~C. Morabito, \enquote{EEG complexity modifications and altered
  compressibility in mild cognitive impairment and Alzheimer’s disease}, {\em
  Recent Advances of Neural Network Models and Applications\/}, pp. 163--173,
  Springer, 2014.

\bibitem{wang2013depression}
Wang, X., C.~Zhang, Y.~Ji, L.~Sun, L.~Wu and Z.~Bao, \enquote{A depression
  detection model based on sentiment analysis in micro-blog social network},
  {\em Pacific-Asia Conference on Knowledge Discovery and Data Mining\/}, pp.
  201--213, Springer, 2013.

\bibitem{pestian2008using}
Pestian, J., P.~Matykiewicz, J.~Grupp-Phelan, S.~A. Lavanier, J.~Combs and
  R.~Kowatch, \enquote{Using natural language processing to classify suicide
  notes}, {\em Proceedings of the Workshop on Current Trends in Biomedical
  Natural Language Processing\/}, pp. 96--97, 2008.

\bibitem{zhou2015tackling}
Zhou, D., J.~Luo, V.~M. Silenzio, Y.~Zhou, J.~Hu, G.~Currier and H.~Kautz,
  \enquote{Tackling mental health by integrating unobtrusive multimodal
  sensing}, {\em Twenty-Ninth AAAI Conference on Artificial Intelligence\/},
  2015.

\bibitem{girard2015automated}
Girard, J.~M. and J.~F. Cohn, \enquote{Automated audiovisual depression
  analysis}, {\em Current Opinion in Psychology\/}, Vol.~4, pp. 75--79, 2015.

\bibitem{meng2013depression}
Meng, H., D.~Huang, H.~Wang, H.~Yang, M.~Ai-Shuraifi and Y.~Wang,
  \enquote{Depression recognition based on dynamic facial and vocal expression
  features using partial least square regression}, {\em Proceedings of the 3rd
  ACM International Workshop on Audio/Visual Emotion Challenge\/}, pp. 21--30,
  2013.

\bibitem{cohn2009detecting}
Cohn, J.~F., T.~S. Kruez, I.~Matthews, Y.~Yang, M.~H. Nguyen, M.~T. Padilla,
  F.~Zhou and F.~De~la Torre, \enquote{Detecting depression from facial actions
  and vocal prosody}, {\em 2009 3rd International Conference on Affective
  Computing and Intelligent Interaction and Workshops\/}, pp. 1--7, IEEE, 2009.

\bibitem{kachele2014fusion}
K{\"a}chele, M., M.~Glodek, D.~Zharkov, S.~Meudt and F.~Schwenker,
  \enquote{Fusion of audio-visual features using hierarchical classifier
  systems for the recognition of affective states and the state of depression},
  {\em Depression\/}, Vol.~1, No.~1, 2014.

\bibitem{karam2014ecologically}
Karam, Z.~N., E.~M. Provost, S.~Singh, J.~Montgomery, C.~Archer, G.~Harrington
  and M.~G. Mcinnis, \enquote{Ecologically valid long-term mood monitoring of
  individuals with bipolar disorder using speech}, {\em 2014 IEEE International
  Conference on Acoustics, Speech and Signal Processing (ICASSP)\/}, pp.
  4858--4862, IEEE, 2014.

\bibitem{eyben2010opensmile}
Eyben, F., M.~W{\"o}llmer and B.~Schuller, \enquote{Opensmile: the munich
  versatile and fast open-source audio feature extractor}, {\em Proceedings of
  the 18th ACM international conference on Multimedia\/}, pp. 1459--1462, ACM,
  2010.

\bibitem{muaremi2014assessing}
Muaremi, A., F.~Gravenhorst, A.~Gr{\"u}nerbl, B.~Arnrich and G.~Tr{\"o}ster,
  \enquote{Assessing bipolar episodes using speech cues derived from phone
  calls}, {\em International Symposium on Pervasive Computing Paradigms for
  Mental Health\/}, pp. 103--114, Springer, 2014.

\bibitem{maxhuni2016classification}
Maxhuni, A., A.~Mu{\~n}oz-Mel{\'e}ndez, V.~Osmani, H.~Perez, O.~Mayora and
  E.~F. Morales, \enquote{Classification of bipolar disorder episodes based on
  analysis of voice and motor activity of patients}, {\em Pervasive and Mobile
  Computing\/}, Vol.~31, pp. 50--66, 2016.

\bibitem{aldeneh2019identifying}
Aldeneh, Z., M.~Jaiswal, M.~Picheny, M.~McInnis and E.~M. Provost,
  \enquote{Identifying Mood Episodes Using Dialogue Features from Clinical
  Interviews}, {\em arXiv preprint arXiv:1910.05115\/}, 2019.

\bibitem{ringeval2018avec}
Ringeval, F., B.~Schuller, M.~Valstar, R.~Cowie, H.~Kaya, M.~Schmitt,
  S.~Amiriparian, N.~Cummins, D.~Lalanne, A.~Michaud {\em et~al.\/},
  \enquote{{AVEC} 2018 workshop and challenge: Bipolar disorder and
  cross-cultural affect recognition}, {\em Proceedings of the 2018 on
  Audio/Visual Emotion Challenge and Workshop\/}, pp. 3--13, ACM, 2018.

\bibitem{yang2018bipolar}
Yang, L., Y.~Li, H.~Chen, D.~Jiang, M.~C. Oveneke and H.~Sahli,
  \enquote{Bipolar Disorder Recognition with Histogram Features of Arousal and
  Body Gestures}, {\em Proceedings of the 2018 on Audio/Visual Emotion
  Challenge and Workshop\/}, pp. 15--21, ACM, 2018.

\bibitem{du2018bipolar}
Du, Z., W.~Li, D.~Huang and Y.~Wang, \enquote{Bipolar Disorder Recognition via
  Multi-scale Discriminative Audio Temporal Representation}, {\em Proceedings
  of the 2018 on Audio/Visual Emotion Challenge and Workshop\/}, pp. 23--30,
  ACM, 2018.

\bibitem{xing2018multi}
Xing, X., B.~Cai, Y.~Zhao, S.~Li, Z.~He and W.~Fan, \enquote{Multi-modality
  Hierarchical Recall based on GBDTs for Bipolar Disorder Classification}, {\em
  Proceedings of the 2018 on Audio/Visual Emotion Challenge and Workshop\/},
  pp. 31--37, ACM, 2018.

\bibitem{syed2018automated}
Syed, Z.~S., K.~Sidorov and D.~Marshall, \enquote{Automated Screening for
  Bipolar Disorder from Audio/Visual Modalities}, {\em Proceedings of the 2018
  on Audio/Visual Emotion Challenge and Workshop\/}, pp. 39--45, ACM, 2018.

\bibitem{ebrahim2018determine}
Ebrahim, M., M.~Al-Ayyoub and M.~Alsmirat, \enquote{Determine Bipolar Disorder
  Level from Patient Interviews Using Bi-LSTM and Feature Fusion}, {\em 2018
  Fifth International Conference on Social Networks Analysis, Management and
  Security (SNAMS)\/}, pp. 182--189, IEEE, 2018.

\bibitem{schuller2019capsule}
{Amiriparian}, S., A.~{Awad}, M.~{Gerczuk}, L.~{Stappen}, A.~{Baird}, S.~{Ottl}
  and B.~{Schuller}, \enquote{Audio-based Recognition of Bipolar Disorder
  Utilising Capsule Networks}, {\em 2019 International Joint Conference on
  Neural Networks\/}, pp. 1--7, July 2019.

\bibitem{ren2019multi}
Ren, Z., J.~Han, N.~Cummins, Q.~Kong, M.~D. Plumbley and B.~W. Schuller,
  \enquote{Multi-instance learning for bipolar disorder diagnosis using weakly
  labelled speech data}, {\em Proceedings of the 9th International Conference
  on Digital Public Health\/}, pp. 79--83, 2019.

\bibitem{zhang2020multimodal}
Zhang, Z., W.~Lin, M.~Liu and M.~Mahmoud, \enquote{Multimodal Deep Learning
  Framework for Mental Disorder Recognition}, {\em 2020 15th IEEE International
  Conference on Automatic Face \& Gesture Recognition (FG 2020). IEEE\/}, 2020.

\bibitem{abaei2020hybrid}
Abaei, N. and H.~Al~Osman, \enquote{A Hybrid Model for Bipolar Disorder
  Classification from Visual Information}, {\em ICASSP 2020-2020 IEEE
  International Conference on Acoustics, Speech and Signal Processing
  (ICASSP)\/}, pp. 4107--4111, IEEE, 2020.

\bibitem{sunbipolar}
Sun, B., S.~Cao, P.~Rao, J.~He, L.~Yu and Y.~Xiao, \enquote{Bipolar Disorder
  Classification Based on Multimodal Recordings}, {\em International Journal of
  Machine Learning and Computing\/}, Vol.~11, No.~1, 2021.

\bibitem{stasak2016investigation}
Stasak, B., J.~Epps, N.~Cummins and R.~Goecke, \enquote{An Investigation of
  Emotional Speech in Depression Classification}, {\em Proc. Interspeech\/},
  pp. 485--489, 2016.

\bibitem{sanchez2013image}
S{\'a}nchez, J., F.~Perronnin, T.~Mensink and J.~Verbeek, \enquote{Image
  classification with the fisher vector: Theory and practice}, {\em
  International Journal of Computer Vision\/}, Vol. 105, No.~3, pp. 222--245,
  2013.

\bibitem{kaya2016fusing}
Kaya, H. and A.~A. Karpov, \enquote{Fusing Acoustic Feature Representations for
  Computational Paralinguistics Tasks.}, {\em Proc. Interspeech\/}, pp.
  2046--2050, 2016.

\bibitem{zong2013weighted}
Zong, W., G.-B. Huang and Y.~Chen, \enquote{Weighted extreme learning machine
  for imbalance learning}, {\em Neurocomputing\/}, Vol. 101, pp. 229--242,
  2013.

\bibitem{goodfellow2013challenges}
Goodfellow, I.~J., D.~Erhan, P.~L. Carrier, A.~Courville, M.~Mirza, B.~Hamner,
  W.~Cukierski, Y.~Tang, D.~Thaler, D.-H. Lee {\em et~al.\/},
  \enquote{Challenges in representation learning: A report on three machine
  learning contests}, {\em International Conference on Neural Information
  Processing\/}, pp. 117--124, Springer, 2013.

\bibitem{sabour2017dynamic}
Sabour, S., N.~Frosst and G.~E. Hinton, \enquote{Dynamic routing between
  capsules}, {\em Advances in Neural Information Processing Systems\/}, pp.
  3856--3866, 2017.

\bibitem{devault2014simsensei}
DeVault, D., R.~Artstein, G.~Benn, T.~Dey, E.~Fast, A.~Gainer, K.~Georgila,
  J.~Gratch, A.~Hartholt, M.~Lhommet {\em et~al.\/}, \enquote{SimSensei Kiosk:
  A virtual human interviewer for healthcare decision support}, {\em
  Proceedings of the 2014 International Conference on Autonomous Agents and
  Multi-Agent Systems\/}, pp. 1061--1068, 2014.

\bibitem{ringeval2019avec}
Ringeval, F., B.~Schuller, M.~Valstar, N.~Cummins, R.~Cowie, L.~Tavabi,
  M.~Schmitt, S.~Alisamir, S.~Amiriparian, E.-M. Messner {\em et~al.\/},
  \enquote{AVEC 2019 workshop and challenge: state-of-mind, detecting
  depression with AI, and cross-cultural affect recognition}, {\em Proceedings
  of the 9th International on Audio/Visual Emotion Challenge and Workshop\/},
  pp. 3--12, 2019.

\bibitem{wang2018deep}
Wang, T., J.~Cao, X.~Lai and B.~Chen, \enquote{Deep weighted extreme learning
  machine}, {\em Cognitive Computation\/}, Vol.~10, No.~6, pp. 890--907, 2018.

\bibitem{yao2018weight}
Yao, P. and X.-Z. Wang, \enquote{Weight Learning in Weighted ELM Classification
  Model Based on Genetic Algorithms}, {\em 2018 International Conference on
  Machine Learning and Cybernetics (ICMLC)\/}, Vol.~2, pp. 370--377, IEEE,
  2018.

\bibitem{young1978rating}
Young, R.~C., J.~T. Biggs, V.~E. Ziegler and D.~A. Meyer, \enquote{A rating
  scale for mania: reliability, validity and sensitivity}, {\em The British
  Journal of Psychiatry\/}, Vol. 133, No.~5, pp. 429--435, 1978.

\bibitem{meral2003analysis}
Meral, H., H.~Ekenel and A.~Ozsoy, \enquote{Analysis of emotion in {Turkish}},
  {\em XVII National Conference on Turkish Linguistics\/}, 2003.

\bibitem{kaya2015contrasting}
Kaya, H., F.~G{\"u}rpinar, S.~Afshar and A.~A. Salah, \enquote{Contrasting and
  combining least squares based learners for emotion recognition in the wild},
  {\em Proceedings of the 2015 ACM on International Conference on Multimodal
  Interaction\/}, pp. 459--466, 2015.

\bibitem{schuller2010interspeech}
Schuller, B., S.~Steidl, A.~Batliner, F.~Burkhardt, L.~Devillers, C.~M{\"u}ller
  and S.~S. Narayanan, \enquote{The {INTERSPEECH} 2010 paralinguistic
  challenge}, {\em Eleventh Annual Conference of the International Speech
  Communication Association\/}, 2010.

\bibitem{eyben2015geneva}
Eyben, F., K.~R. Scherer, B.~W. Schuller, J.~Sundberg, E.~Andr{\'e}, C.~Busso,
  L.~Y. Devillers, J.~Epps, P.~Laukka, S.~S. Narayanan {\em et~al.\/},
  \enquote{The Geneva minimalistic acoustic parameter set (GeMAPS) for voice
  research and affective computing}, {\em IEEE Transactions on Affective
  Computing\/}, Vol.~7, No.~2, pp. 190--202, 2015.

\bibitem{hilty1999review}
Hilty, D.~M., K.~T. Brady and R.~E. Hales, \enquote{A review of bipolar
  disorder among adults}, {\em Psychiatric Services\/}, Vol.~50, No.~2, pp.
  201--213, 1999.

\bibitem{radford2019language}
Radford, A., J.~Wu, R.~Child, D.~Luan, D.~Amodei and I.~Sutskever,
  \enquote{Language models are unsupervised multitask learners}, {\em OpenAI
  Blog\/}, Vol.~1, No.~8, p.~9, 2019.

\bibitem{devlin2018bert}
Devlin, J., M.-W. Chang, K.~Lee and K.~Toutanova, \enquote{Bert: Pre-training
  of deep bidirectional transformers for language understanding}, {\em arXiv
  preprint arXiv:1810.04805\/}, 2018.

\bibitem{brown2020language}
Brown, T.~B., B.~Mann, N.~Ryder, M.~Subbiah, J.~Kaplan, P.~Dhariwal,
  A.~Neelakantan, P.~Shyam, G.~Sastry, A.~Askell {\em et~al.\/},
  \enquote{Language models are few-shot learners}, {\em arXiv preprint
  arXiv:2005.14165\/}, 2020.

\bibitem{pennebaker2001linguistic}
Pennebaker, J., L.~Francis and R.~Booth, \enquote{Linguistic inquiry and word
  count: LIWC2001}, {\em LIWC Operators Manual 2001\/}, 2001.

\bibitem{trstenjak2014knn}
Trstenjak, B., S.~Mikac and D.~Donko, \enquote{KNN with TF-IDF based framework
  for text categorization}, {\em Procedia Engineering\/}, Vol.~69, pp.
  1356--1364, 2014.

\bibitem{van2010automatic}
Van~Zaanen, M. and P.~Kanters, \enquote{Automatic Mood Classification Using TF*
  IDF Based on Lyrics.}, {\em ISMIR\/}, pp. 75--80, 2010.

\bibitem{hiemstra2000probabilistic}
Hiemstra, D., \enquote{A probabilistic justification for using tf$\times$ idf
  term weighting in information retrieval}, {\em International Journal on
  Digital Libraries\/}, Vol.~3, No.~2, pp. 131--139, 2000.

\bibitem{jing2002improved}
Jing, L.-P., H.-K. Huang and H.-B. Shi, \enquote{Improved feature selection
  approach TFIDF in text mining}, {\em Proceedings. International Conference on
  Machine Learning and Cybernetics\/}, Vol.~2, pp. 944--946, IEEE, 2002.

\bibitem{bird2009natural}
Bird, S., E.~Klein and E.~Loper, {\em Natural language processing with Python:
  analyzing text with the natural language toolkit\/}, O'Reilly Media, Inc.,
  2009.

\bibitem{porter1980algorithm}
Porter, M.~F., \enquote{An algorithm for suffix stripping}, {\em Program\/},
  Vol.~14, No.~3, pp. 130--137, 1980.

\bibitem{gilbert2014vader}
Gilbert, C. and E.~Hutto, \enquote{Vader: A parsimonious rule-based model for
  sentiment analysis of social media text}, {\em Eighth International
  Conference on Weblogs and Social Media (ICWSM-14). Available at (20/04/16)
  http://comp. social. gatech. edu/papers/icwsm14. vader. hutto. pdf\/},
  Vol.~81, p.~82, 2014.

\bibitem{loria2018textblob}
Loria, S., \enquote{textblob Documentation}, {\em Release 0.15\/}, Vol.~2,
  2018.

\bibitem{akbik2019flair}
Akbik, A., T.~Bergmann, D.~Blythe, K.~Rasul, S.~Schweter and R.~Vollgraf,
  \enquote{Flair: An easy-to-use framework for state-of-the-art nlp}, {\em
  Proceedings of the 2019 Conference of the North American Chapter of the
  Association for Computational Linguistics (Demonstrations)\/}, pp. 54--59,
  2019.

\bibitem{kaya2017video}
Kaya, H., F.~G{\"u}rp{\i}nar and A.~A. Salah, \enquote{Video-based emotion
  recognition in the wild using deep transfer learning and score fusion}, {\em
  Image and Vision Computing\/}, Vol.~65, pp. 66--75, 2017.

\bibitem{huang2011extreme}
Huang, G.-B., H.~Zhou, X.~Ding and R.~Zhang, \enquote{Extreme learning machine
  for regression and multiclass classification}, {\em IEEE Transactions on
  Systems, Man, and Cybernetics, Part B (Cybernetics)\/}, Vol.~42, No.~2, pp.
  513--529, 2011.

\bibitem{rao1972generalized}
Karantha, M.~P., S.~Sheela and D.~Purushothama, \enquote{Generalized inverse of
  a matrix and its applications}, {\em Proceedings of the Sixth Berkeley
  Symposium on Mathematical Statistics and Probability, Volume 1: Theory of
  Statistics\/}, The Regents of the University of California, 1972.

\bibitem{gurpinar2016kernel}
Gurpinar, F., H.~Kaya, H.~Dibeklioglu and A.~Salah, \enquote{Kernel ELM and CNN
  based facial age estimation}, {\em Proceedings of the IEEE conference on
  computer vision and pattern recognition workshops\/}, pp. 80--86, 2016.

\bibitem{d2015review}
D'mello, S.~K. and J.~Kory, \enquote{A review and meta-analysis of multimodal
  affect detection systems}, {\em ACM Computing Surveys (CSUR)\/}, Vol.~47,
  No.~3, pp. 1--36, 2015.

\bibitem{gunes2008automatic}
Gunes, H. and M.~Piccardi, \enquote{Automatic temporal segment detection and
  affect recognition from face and body display}, {\em IEEE Transactions on
  Systems, Man, and Cybernetics, Part B (Cybernetics)\/}, Vol.~39, No.~1, pp.
  64--84, 2008.

\end{thebibliography}

\end{document}